# Multilingual Part-of-Speech Tagging: Two Unsupervised Approaches

**Tahira Naseem**                                    TAHIRA@CSAIL.MIT.EDU
**Benjamin Snyder**                                  BSNYDER@CSAIL.MIT.EDU
**Jacob Eisenstein**                                 JACOBE@CSAIL.MIT.EDU
**Regina Barzilay**                                  REGINA@CSAIL.MIT.EDU
*Computer Science and Artificial Intelligence Laboratory*
*Massachusetts Institute of Technology*
*77 Massachusetts Avenue, Cambridge MA 02139*

## Abstract

We demonstrate the effectiveness of multilingual learning for unsupervised part-of-speech tagging. The central assumption of our work is that by combining cues from multiple languages, the structure of each becomes more apparent. We consider two ways of applying this intuition to the problem of unsupervised part-of-speech tagging: a model that directly merges tag structures for a pair of languages into a single sequence and a second model which instead incorporates multilingual context using latent variables. Both approaches are formulated as hierarchical Bayesian models, using Markov Chain Monte Carlo sampling techniques for inference. Our results demonstrate that by incorporating multilingual evidence we can achieve impressive performance gains across a range of scenarios. We also found that performance improves steadily as the number of available languages increases.

## 1. Introduction

In this paper, we explore the application of multilingual learning to part-of-speech tagging when no annotation is available.[1] The fundamental idea upon which our work is based is that the patterns of ambiguity inherent in part-of-speech tag assignments differ across languages. At the lexical level, a word with part-of-speech tag ambiguity in one language may correspond to an unambiguous word in the other language. For example, the word "can" in English may function as an auxiliary verb, a noun, or a regular verb. However, many other languages are likely to express these different senses with three distinct lexemes. Languages also differ in their patterns of structural ambiguity. For example, the presence of an article in English greatly reduces the ambiguity of the succeeding tag. In languages without articles, however, this constraint is obviously absent. The key idea of multilingual learning is that by combining natural cues from multiple languages, the structure of each becomes more apparent.

Even in expressing the same meaning, languages take different syntactic routes, leading to cross-lingual variation in part-of-speech patterns. Therefore, an effective multilingual model must accurately represent common linguistic structure, yet remain flexible to the idiosyncrasies of each language. This tension only becomes stronger as additional languages are added to the mix. Thus, a key challenge of multilingual learning is to capture cross-lingual correlations while preserving individual language tagsets, tag selections, and tag orderings.

---

1. Code, data sets, and the raw outputs of our experiments are available at http://groups.csail.mit.edu/rbg/code/multiling_pos.

In this paper, we explore two different approaches for modeling cross-lingual correlations. The first approach directly merges pairs of tag sequences into a single bilingual sequence, employing joint distributions over aligned tag-pairs; for unaligned tags, language-specific distributions are still used. The second approach models multilingual context using latent variables instead of explicit node merging. For a group of aligned words, the multilingual context is encapsulated in the value of a corresponding latent variable. Conditioned on the latent variable, the tagging decisions for each language remain independent. In contrast to the first model, the architecture of the hidden variable model allows it to scale gracefully as the number of languages increases.

Both approaches are formulated as hierarchical Bayesian models with an underlying trigram HMM substructure for each language. The first model operates as a simple directed graphical model with only one additional *coupling* parameter beyond the transition and emission parameters used in monolingual HMMs. The latent variable model, on the other hand, is formulated as a non-parametric model; it can be viewed as performing multilingual clustering on aligned sets of tag variables. Each latent variable value indexes a separate distribution on tags for each language, appropriate to the given context. For both models, we perform inference using Markov Chain Monte Carlo sampling techniques.

We evaluate our models on a parallel corpus of eight languages: Bulgarian, Czech, English, Estonian, Hungarian, Romanian, Serbian, and Slovene. We consider a range of scenarios that vary from combinations of bilingual models to a single model that is jointly trained on all eight languages. Our results show consistent and robust improvements over a monolingual baseline for almost all combinations of languages. When a complete tag lexicon is available and the latent variable model is trained using eight languages, average performance increases from 91.1% accuracy to 95%, more than halving the gap between unsupervised and supervised performance. In more realistic cases, where the lexicon is restricted to only frequently occurring words, we see even larger gaps between monolingual and multilingual performance. In one such scenario, average multilingual performance increases to 82.8% from a monolingual baseline of 74.8%. For some language pairs, the improvement is especially noteworthy; for instance, in complete lexicon scenario, Serbian improves from 84.5% to 94.5% when paired with English.

We find that in most scenarios the latent variable model achieves higher performance than the merged structure model, even when it too is restricted to pairs of languages. Moreover the hidden variable model can effectively accommodate large numbers of languages which makes it a more desirable framework for multilingual learning. However, we observe that the latent variable model is somewhat sensitive to lexicon coverage. The performance of the merged structure model, on the other hand, is more robust in this respect. In the case of the drastically reduced lexicon (with 100 words only), its performance is clearly better than the hidden variable model. This indicates that the merged structure model might be a better choice for the languages that lack lexicon resources.

A surprising discovery of our experiments is the marked variation in the level of improvement across language pairs. If the best pairing for each language is chosen by an oracle, average bilingual performance reaches 95.4%, compared to average performance of 93.1% across all pairs. Our experiments demonstrate that this variability is influenced by cross-lingual links between languages as well as by the model under consideration. We identify several factors that contribute to the success of language pairings, but none of them can uniquely predict which supplementary language is most helpful. These results suggest that when multi-parallel corpora are available, a model that simultaneously exploits all the languages – such as the latent variable model proposed here – is preferable to a strategy that selects one of the bilingual models. We found that performance tends to improves steadily as the number of available languages increases.

In realistic scenarios, tagging resources for some number of languages may already be available. Our models can easily exploit any amount of tagged data in any subset of available languages. As our experiments show, as annotation is added, performance increases even for those languages lacking resources.

The remainder of the paper is structured as follows. Section 2 compares our approach with previous work on multilingual learning and unsupervised part-of-speech tagging. Section 3 presents two approaches for modeling multilingual tag sequences, along with their inference procedures and implementation details. Section 4 describes corpora used in the experiments, preprocessing steps and various evaluation scenarios. The results of the experiments and their analysis are given in Sections 5, and 6. We summarize our contributions and consider directions for future work in Section 7.

## 2. Related Work

We identify two broad areas of related work: multilingual learning and inducing part-of-speech tags without labeled data. Our discussion of multilingual learning focuses on unsupervised approaches that incorporate two or more languages. We then describe related work on unsupervised and semi-supervised models for part-of-speech tagging.

### 2.1 Multilingual Learning

The potential of multilingual data as a rich source of linguistic knowledge has been recognized since the early days of empirical natural language processing. Because patterns of ambiguity vary greatly across languages, unannotated multilingual data can serve as a learning signal in an unsupervised setting. We are especially interested in methods to leverage more than two languages jointly, and compare our approach with relevant prior work.

Multilingual learning may also be applied in a semi-supervised setting, typically by projecting annotations across a parallel corpus to another language where such resources do not exist (e.g., Yarowsky, Ngai, & Wicentowski, 2000; Diab & Resnik, 2002; Padó & Lapata, 2006; Xi & Hwa, 2005). As our primary focus is on the unsupervised induction of cross-linguistic structures, we do not address this area.

#### 2.1.1 Bilingual Learning

Word sense disambiguation (WSD) was among the first successful applications of automated multilingual learning (Dagan et al., 1991; Brown et al., 1991). Lexical ambiguity differs across languages – each sense of a polysemous word in one language may translate to a distinct counterpart in another language. This makes it possible to use aligned foreign-language words as a source of noisy supervision. Bilingual data has been leveraged in this way in a variety of WSD models (Brown et al., 1991; Resnik & Yarowsky, 1997; Ng, Wang, & Chan, 2003; Diab & Resnik, 2002; Li & Li, 2002; Bhattacharya, Getoor, & Bengio, 2004), and the quality of supervision provided by multilingual data closely approximates that of manual annotation (Ng et al., 2003). Polysemy is one source of ambiguity for part-of-speech tagging; thus our model implicitly leverages multilingual WSD in the context of a higher-level syntactic analysis.

Multilingual learning has previously been applied to syntactic analysis; a pioneering effort was the inversion transduction grammar of Wu (1995). This method is trained on an unannotated parallel corpus using a probabilistic bilingual lexicon and deterministic constraints on bilingual tree structures. The inside-outside algorithm (Baker, 1979) is used to learn parameters for manually specified bilingual grammar. These ideas were extended by subsequent work on synchronous grammar induction and hierarchical phrase-based translation (Wu & Wong, 1998; Chiang, 2005).

One characteristic of this family of methods is that they were designed for inherently multilingual tasks such as machine translation and lexicon induction. While we share the goal of *learning* from multilingual data, we seek to induce monolingual syntactic structures that can be applied even when multilingual data is unavailable.

In this respect, our approach is closer to the unsupervised multilingual grammar induction work of Kuhn (2004). Starting from the hypothesis that trees induced over parallel sentences should exhibit cross-lingual structural similarities, Kuhn uses word-level alignments to constrain the set of plausible syntactic constituents. These constraints are implemented through hand-crafted deterministic rules, and are incorporated in expectation-maximization grammar induction to assign zero likelihood to illegal bracketings. The probabilities of the productions are then estimated separately for each language, and can be applied to monolingual data directly. Kuhn shows that this form of multilingual training yields better monolingual parsing performance.

Our methods incorporate cross-lingual information in a fundamentally different manner. Rather than using hand-crafted deterministic rules – which may require modification for each language pair – we estimate probabilistic multilingual patterns directly from data. Moreover, the estimation of multilingual patterns is incorporated directly into the tagging model itself.

Finally, multilingual learning has recently been applied to unsupervised morphological segmentation (Snyder & Barzilay, 2008). This research is related, but moving from morphological to syntactic analysis imposes new challenges. One key difference is that Snyder & Barzilay model morphemes as unigrams, ignoring the transitions between morphemes. In syntactic analysis, transition information provides a crucial constraint, requiring a fundamentally different model structure.

### 2.1.2 BEYOND BILINGUAL LEARNING

While most work on multilingual learning focuses on bilingual analysis, some models operate on more than one pair of languages. For instance, Genzel (2005) describes a method for inducing a multilingual lexicon from a group of related languages. This work first induces bilingual models for each pair of languages and then combines them. We take a different approach by simultaneously learning from all languages, rather than combining bilingual results.

A related thread of research is multi-source machine translation (Och & Ney, 2001; Utiyama & Isahara, 2006; Cohn & Lapata, 2007; Chen, Eisele, & Kay, 2008; Bertoldi, Barbaiani, Federico, & Cattoni, 2008) where the goal is to translate from multiple source languages to a single target language. By using multi-source corpora, these systems alleviate sparseness and increase translation coverage, thereby improving overall translation accuracy. Typically, multi-source translation systems build separate bilingual models and then select a final translation from their output. For instance, a method developed by Och and Ney (2001) generates several alternative translations from source sentences expressed in different languages and selects the most likely candidate. Cohn and Lapata (2007) consider a different generative model: rather than combining alternative sentence translations in a post-processing step, their model estimates the target phrase translation distribu-tion by marginalizing over multiple translations from various source languages. While their model combines multilingual information at the phrase level, at its core are estimates for phrase tables that are obtained using bilingual models.

In contrast, we present an approach for unsupervised multilingual learning that builds a single joint model across all languages. This makes maximal use of unlabeled data and sidesteps the difficult problem of combining the output of multiple bilingual systems without supervision.

## 2.2 Unsupervised Part-of-Speech Tagging

Unsupervised part-of-speech tagging involves predicting the tags for words, without annotations of the correct tags for any word tokens. Generally speaking, the unsupervised setting does permit the use of declarative knowledge about the relationship between tags and word *types*, in the form of a dictionary of the permissible tags for the most common words. This setup is referred to as "semi-supervised" by Toutanova and Johnson (2008), but is considered "unsupervised" in most other papers on the topic (e.g., Goldwater & Griffiths, 2007). Our evaluation considers tag dictionaries of varying levels of coverage.

Since the work of Merialdo (1994), the hidden Markov model (HMM) has been the most common representation[2] for unsupervised tagging (Banko & Moore, 2004). Part-of-speech tags are encoded as a linear chain of hidden variables, and words are treated as emitted observations. Recent advances include the use of a fully Bayesian HMM (Johnson, 2007; Goldwater & Griffiths, 2007), which places prior distributions on tag transition and word-emission probabilities. Such Bayesian priors permit integration over parameter settings, yielding models that perform well across a range of settings. This is particularly important in the case of small datasets, where many of the counts used for maximum-likelihood parameter estimation will be sparse. The Bayesian setting also facilitates the integration of other data sources, and thus serves as the departure point for our work.

Several recent papers have explored the development of alternative training procedures and model structures in an effort to incorporate more expressive features than permitted by the generative HMM. Smith and Eisner (2005) maintain the HMM structure, but incorporate a large number of overlapping features in a conditional log-linear formulation. *Contrastive estimation* is used to provide a training criterion which maximizes the probability of the observed sentences compared to a set of similar sentences created by perturbing word order. The use of a large set of features and a discriminative training procedure led to strong performance gains.

Toutanova and Johnson (2008) propose an LDA-style model for unsupervised part-of-speech tagging, grouping words through a latent layer of ambiguity classes. Each ambiguity class corresponds to a set of permissible tags; in many languages this set is tightly constrained by morphological features, thus allowing an incomplete tagging lexicon to be expanded. Haghighi and Klein (2006) also use a variety of morphological features, learning in an undirected Markov Random Field that permits overlapping features. They propagate information from a small number of labeled "prototype" examples using the distributional similarity between prototype and non-prototype words.

Our focus is to effectively incorporate multilingual evidence, and we require a simple model that can easily be applied to multiple languages with widely varying structural properties. We view this direction as orthogonal to refining monolingual tagging models for any particular language.

---

2. In addition to the basic HMM architecture, other part-of-speech tagging approaches have been explored (Brill, 1995; Mihalcea, 2004)

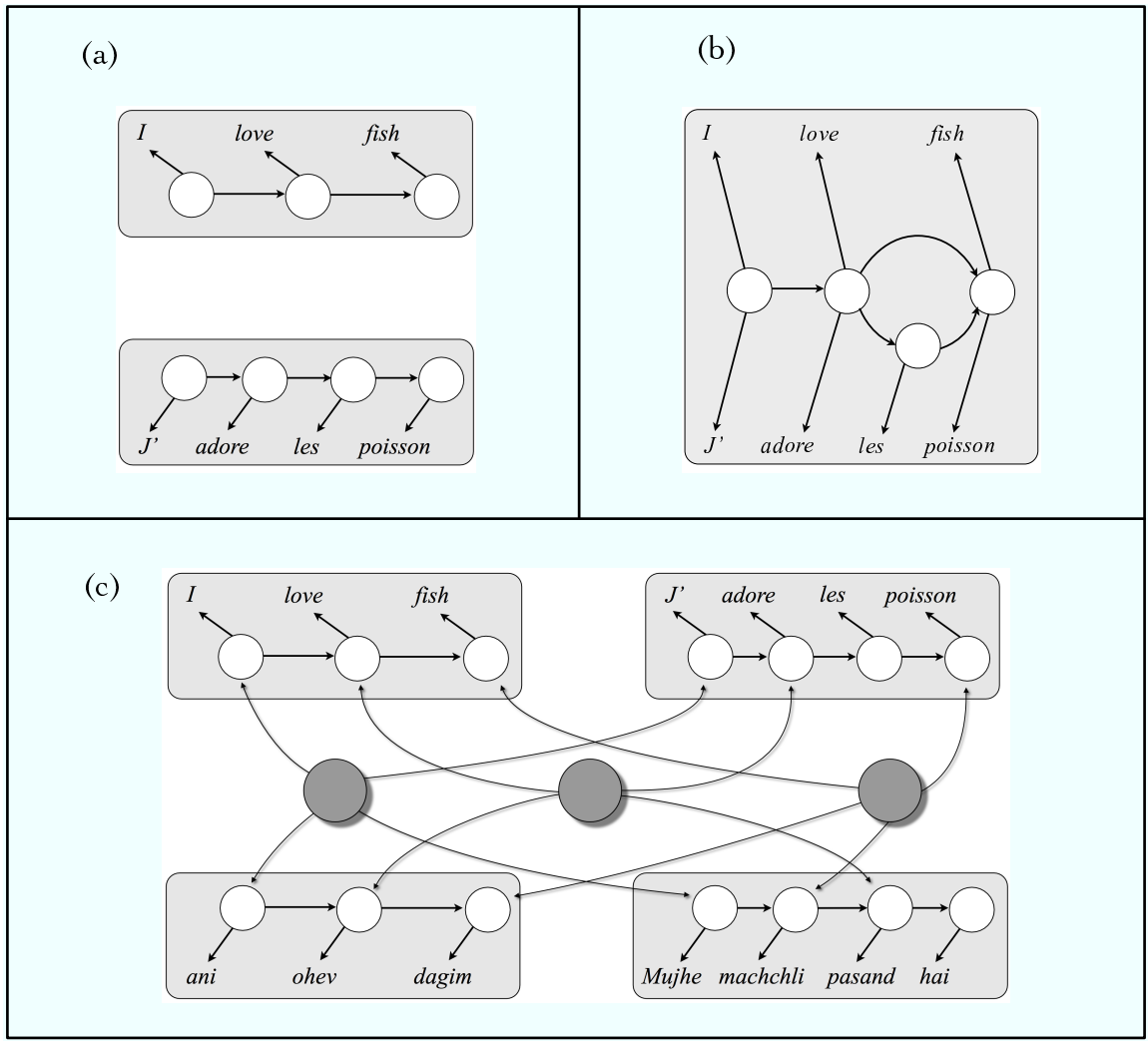

Figure 1: Example graphical structures of (a) two standard monolingual HMMs, (b) our merged node model, and (c) our latent variable model with three superlingual variables.

## 3. Models

The motivating hypothesis of this work is that patterns of ambiguity at the part-of-speech level differ across languages in systematic ways. By considering multiple languages simultaneously, the total inherent ambiguity can be reduced in each language. But with the potential advantages of leveraging multilingual information comes the challenge of respecting language-specific characteristics such as tag inventory, selection and order. To this end, we develop models that jointly tag parallel streams of text in multiple languages, while maintaining language-specific tag sets and parameters over transitions and emissions.

Part-of-speech tags reflect the syntactic and semantic function of the tagged words. Across languages, pairs of word tokens that are known to share semantic or syntactic function should have tags that are related in systematic ways. The *word alignment* task in machine translation is to identify just such pairs of words in parallel sentences. Aligned word pairs serve as the cross-lingual anchors of our model, allowing information to be shared via joint tagging decisions. Research in machine translation has produced robust tools for identifying word alignments; we use such a tool as a black box and treat its output as a fixed, observed property of the parallel data.

Given a set of parallel sentences, we posit a hidden Markov model (HMM) for each language, where the hidden states represent the tags and the emissions are the words. In the unsupervised monolingual setting, inference on the part-of-speech tags is performed jointly with estimation of parameters governing the relationship between tags and words (the *emission* probabilities) and between consecutive tags (the *transition* probabilities). Our multilingual models are built upon this same structural foundation, so that the emission and transition parameters retain an identical interpretation as in the monolingual setting. Thus, these parameters can be learned on parallel text and later applied to monolingual data.

We consider two alternative approaches for incorporating cross-lingual information. In the first model, the tags for aligned words are merged into single bi-tag nodes; in the second, latent variable model, an additional layer of hidden *superlingual tags* instead exerts influence on the tags of clusters of aligned words. The first model is primarily designed for bilingual data, while the second model operates over any number of languages. Figure 1 provides a graphical model representation of the monolingual, merged node, and latent variable models instantiated over a single parallel sentence.

Both the merged node and latent variable approaches are formalized as hierarchical Bayesian models. This provides a principled probabilistic framework for integrating multiple sources of information, and offers well-studied inference techniques. Table 1 summarizes the mathematical notation used throughout this section. We now describe each model in depth.

### 3.1 Bilingual Unsupervised Tagging: A Merged Node Model

In the bilingual merged node model, cross-lingual context is incorporated by creating joint bi-tag nodes for aligned words. It would be too strong to insist that aligned words have an identical tag; indeed, it may not even be appropriate to assume that two languages share identical tag sets. However, when two words are aligned, we do want to choose their tags jointly. To enable this, we allow the values of the bi-tag nodes to range over all possible tag pairs $\langle t, t' \rangle \in T \times T'$, where $T$ and $T'$ represent the tagsets for each language.

The tags $t$ and $t'$ need not be identical, but we do believe that they are systematically related. This is modeled using a coupling distribution $\omega$, which is multinomial over all tag pairs. The parameter $\omega$ is combined with the standard transition distribution $\phi$ in a product-of-experts model. Thus, the aligned tag pair $\langle y_i, y'_j \rangle$ is conditioned on the predecessors $y_{i-1}$ and $y'_{j-1}$, as well as the coupling parameter $\omega(y_i, y'_j)$.[3] The coupled bi-tag nodes serve as bilingual "anchors" – due to the Markov dependency structure, even unaligned words may benefit from cross-lingual information that propagates from these nodes.

---

3. While describing the merged node model, we consider only the two languages $\ell$ and $\ell'$, and use a simplified notation in which we write $\langle y, y' \rangle$ to mean $\langle y^\ell, y^{\ell'} \rangle$. Similar abbreviations are used for the language-indexed parameters.

**Notation used in both models**

| | | |
|---|---|---|
| $x_i^\ell$ | – | The $i^{th}$ word token of the sentence in language $\ell$. |
| $y_i^\ell$ | – | The $i^{th}$ tag of the sentence in language $\ell$. |
| $\mathbf{a}^{\ell,\ell'}$ | – | The word alignments for the language pair $\langle \ell, \ell' \rangle$. |
| $\phi_t^\ell$ | – | The transition distribution (over tags), conditioned on tag $t$ in language $\ell$. We describe a bigram transition model, though our implementation uses trigrams (without bigram interpolations); the extension is trivial. |
| $\theta_t^\ell$ | – | The emission distribution (over words), conditioned on tag $t$ in language $\ell$. |
| $\phi_0$ | – | The parameter of the symmetric Dirichlet prior on the transition distributions. |
| $\theta_0$ | – | The parameter of the symmetric Dirichlet prior on the emission distributions. |

**Notation used in the merged node model**

| | | |
|---|---|---|
| $\omega$ | – | A coupling parameter that assigns probability mass to each pair of aligned tags. |
| $\omega_0$ | – | A Dirichlet prior on the coupling parameter. |
| $A_b$ | – | Distribution over bilingual alignments. |

**Notation used in the latent variable model**

| | | |
|---|---|---|
| $\pi$ | – | A multinomial over the superlingual tags $\mathbf{z}$. |
| $\alpha$ | – | The *concentration parameter* for $\pi$, controlling how much probability mass is allocated to the first few values. |
| $z_j$ | – | The setting of the $j^{th}$ superlingual tag, ranging over the set of integers, and indexing a distribution set in $\Psi$. |
| $\Psi_z = \langle \psi_z^1, \psi_z^2, \ldots, \psi_z^n \rangle$ | – | The $z^{th}$ set of distributions over tags in all languages $\ell_1$ through $\ell_n$. |
| $G_0$ | – | A base distribution from which the $\Psi_z$ are drawn, whose form is a set of $n$ symmetric Dirichlet distributions each with a parameter $\psi_0$. |
| $A_m$ | – | Distribution over multilingual alignments. |

Table 1: Summary of notation used in the description of both models. As each sentence is treated in isolation (conditioned on the parameters), the sentence indexing is left implicit.

We now present a generative account of how the words in each sentence and the parameters of the model are produced. This generative story forms the basis of our sampling-based inference procedure.

### 3.1.1 Merged Node Model: Generative Story

Our generative story assumes the existence of two tagsets, $T$ and $T'$, and two vocabularies $W$ and $W'$ – one of each for each language. For ease of exposition, we formulate our model with bigram tag dependencies. However, in our experiments we used a trigram model (without any bigram interpolation), which is a trivial extension of the described model.

1. **Transition and Emission Parameters**. For each tag $t \in T$, draw a *transition* distribution $\phi_t$ over tags $T$, and an *emission* distribution $\theta_t$ over words $W$. Both the transition and emission distributions are multinomial, so they are drawn from their conjugate prior, the Dirichlet (Gelman, Carlin, Stern, & Rubin, 2004). We use symmetric Dirichlet priors, which encode only an expectation about the uniformity of the induced multinomials, but not do encode preferences for specific words or tags.

   For each tag $t \in T'$, draw a *transition* distribution $\phi'_t$ over tags $T'$, and an *emission* distribution $\theta'_t$ over words $W'$, both from symmetric Dirichlet priors.

2. **Coupling Parameter**. Draw a bilingual *coupling* distribution $\omega$ over tag pairs pairs $T \times T'$. This distribution is multinomial with dimension $|T| \cdot |T'|$, and is drawn from a symmetric Dirichlet prior $\omega_0$ over all tag pairs.

3. **Data**. For each bilingual parallel sentence:

   (a) Draw an alignment $\mathbf{a}$ from a bilingual alignment distribution $A_b$. The following paragraph defines $\mathbf{a}$ and $A_b$ more formally.

   (b) Draw a bilingual sequence of part-of-speech tags $(y_1, ..., y_m)$, $(y'_1, ..., y'_n)$ according to: $P((y_1, ..., y_m), (y'_1, ..., y'_n)|\mathbf{a}, \phi, \phi', \omega)$.[4] This joint distribution thus conditions on the alignment structure, the transition probabilities for both languages, and the coupling distribution; a formal definition is given in Formula 1.

   (c) For each part-of-speech tag $y_i$ in the first language, emit a word from the vocabulary $W$: $x_i \sim \theta_{y_i}$,

   (d) For each part-of-speech tag $y'_j$ in the second language, emit a word from the vocabulary $W'$: $x'_j \sim \theta'_{y'_j}$.

This completes the outline of the generative story. We now provide more detail on how alignments are handled, and on the distribution over coupled part-of-speech tag sequences.

**Alignments**    An alignment $\mathbf{a}$ defines a bipartite graph between the words $\mathbf{x}$ and $\mathbf{x}'$ in two parallel sentences . In particular, we represent $\mathbf{a}$ as a set of integer pairs, indicating the word indices. Crossing edges are not permitted, as these would lead to cycles in the resulting graphical model; thus, the existence of an edge $(i, j)$ precludes any additional edges $(i + a, j - b)$ or $(i - a, j + b)$, for $a, b \geq 0$. From a linguistic perspective, we assume that the edge $(i, j)$ indicates that the words $x_i$ and $x'_j$ share some syntactic and/or semantic role in the bilingual parallel sentences.

---

4. We use a special end state, rather than explicitly modeling sentence length. Thus the values of $m$ and $n$ are determined stochastically.

From the perspective of the generative story, alignments are treated as draws from a distribution $A_b$. Since the alignments are always observed, we can remain agnostic about the distribution $A_b$, except to require that it assign zero probability to alignments which either: *(i)* align a single index in one language to multiple indices in the other language or *(ii)* contain crossing edges. The resulting alignments are thus one-to-one, contain no crossing edges, and may be sparse or even possibly empty. Our technique for obtaining alignments that display these properties is described in Section 4.2.

**Generating Tag Sequences**   In a standard hidden Markov model for part-of-speech tagging, the tags are drawn as a Markov process from the transition distribution. This permits the probability of a tag sequence to factor across the time steps. Our model employs a similar factorization: the tags for unaligned words are drawn from their predecessor's transition distribution, while joined tag nodes are drawn from a product involving the coupling parameter and the transition distributions for both languages.

More formally, given an alignment $\mathbf{a}$ and sets of transition parameters $\phi$ and $\phi'$, we factor the conditional probability of a bilingual tag sequence $(y_1, ..., y_m)$, $(y'_1, ..., y'_n)$ into transition probabilities for unaligned tags, and joint probabilities over aligned tag pairs:

$$P((y_1, ..., y_m), (y'_1, ..., y'_n)|\mathbf{a}, \phi, \phi', \omega) = \prod_{\text{unaligned } i} \phi_{y_{i-1}}(y_i) \prod_{\text{unaligned } j} \phi'_{y'_{j-1}}(y'_j)$$

$$\prod_{(i,j) \in \mathbf{a}} P(y_i, y'_j | y_{i-1}, y'_{j-1}, \phi, \phi', \omega). \qquad (1)$$

Because the alignment contains no crossing edges, we can still model the tags as generated sequentially by a stochastic process. We define the distribution over aligned tag pairs to be a product of each language's transition probability and the coupling probability:

$$P(y_i, y'_j | y_{i-1}, y'_{j-1}, \phi, \phi', \omega) = \frac{\phi_{y_{i-1}}(y_i) \; \phi'_{y'_{j-1}}(y'_j) \omega(y_i, y'_j)}{Z}. \qquad (2)$$

The normalization constant here is defined as:

$$Z = \sum_{y, y'} \phi_{y_{i-1}}(y) \; \phi'_{y'_{j-1}}(y') \; \omega(y, y').$$

This factorization allows the language-specific transition probabilities to be shared across aligned and unaligned tags.

Another way to view this probability distribution is as a product of three experts: the two transition parameters and the coupling parameter. Product-of-expert models (Hinton, 1999) allow each information source to exercise very strong negative influence on the probability of tags that they consider to be inappropriate, as compared with additive models. This is ideal for our setting, as it prevents the coupling distribution from causing the model to generate a tag that is unacceptable from the perspective of the monolingual transition distribution. In preliminary experiments we found that a multiplicative approach was strongly preferable to additive models.

### 3.1.2 Merged Node Model: Inference

The goal of our inference procedure is to obtain transition and emission parameters $\theta$ and $\phi$ that can be applied to monolingual test data. Ideally we would choose the parameters that have the highest marginal probability, conditioned on the observed words $\mathbf{x}$ and alignments $\mathbf{a}$,

$$\hat{\theta}, \hat{\phi} = \arg \max_{\theta, \phi} \int P(\theta, \phi, \mathbf{y}, \omega | \mathbf{x}, \mathbf{a}, \theta_0, \phi_0, \omega_0) d\mathbf{y} d\omega.$$

While the structure of our model permits us to decompose the joint probability, it is not possible to analytically marginalize all of the hidden variables. We resort to standard Monte Carlo approximation, in which marginalization is performed through sampling. By repeatedly sampling individual hidden variables according to the appropriate distributions, we obtain a Markov chain that is guaranteed to converge to a stationary distribution centered on the desired posterior. Thus, after an initial burn-in phase, we can use the samples to approximate a marginal distribution over any desired parameter (Gilks, Richardson, & Spiegelhalter, 1996).

The core element of our inference procedure is Gibbs sampling (Geman & Geman, 1984). Gibbs sampling begins by randomly initializing all unobserved random variables; at each iteration, each random variable $u_i$ is then sampled from the conditional distribution $P(u_i | \mathbf{u}_{-i})$, where $\mathbf{u}_{-i}$ refers to all variables other than $u_i$. Eventually, the distribution over samples drawn from this process will converge to the unconditional joint distribution $P(\mathbf{u})$ of the unobserved variables. When possible, we avoid explicitly sampling variables which are not of direct interest, but rather integrate over them. This technique is known as *collapsed sampling*; it is guaranteed never to increase sampling variance, and will often reduce it (Liu, 1994).

In the merged node model, we need sample only the part-of-speech tags and the priors. We are able to exactly marginalize the emission parameters $\theta$ and approximately marginalize the transition and coupling parameters $\phi$ and $\omega$ (the approximations are required due to the re-normalized product of experts — see below for more details). We draw repeated samples of the part-of-speech tags, and construct a sample-based estimate of the underlying tag sequence. After sampling, we construct maximum *a posteriori* estimates of the parameters of interest for each language, $\theta$ and $\phi$.

**Sampling Unaligned Tags** For unaligned part-of-speech tags, the conditional sampling equations are identical to the monolingual Bayesian hidden Markov model. The probability of each tag decomposes into three factors:

$$P(y_i | \mathbf{y}_{-i}, \mathbf{y}', \mathbf{x}, \mathbf{x}', \theta_0, \phi_0) \propto P(x_i | y_i, \mathbf{y}_{-i}, \mathbf{x}_{-i}, \theta_0) P(y_i | y_{i-1}, \mathbf{y}_{-i}, \phi_0) P(y_{i+1} | y_i, \mathbf{y}_{-i}, \phi_0), \quad (3)$$

which follows from the chain rule and the conditional independencies in the model. The first factor is for the emission $x_i$ and the remaining two are for the transitions. We now derive the form of each factor, marginalizing the parameters $\theta$ and $\phi$.

For the emission factor, we can exactly marginalize the emission distribution $\theta$, whose prior is Dirichlet with hyperparameter $\theta_0$. The resulting distribution is a ratio of counts, where the prior acts as a pseudo-count:

$$P(x_i | \mathbf{y}, \mathbf{x}_{-i}, \theta_0) = \int_{\theta_{y_i}} \theta_{y_i}(x_i) P(\theta_{y_i} | \mathbf{y}, \mathbf{x}_{-i}, \theta_0) d\theta_{y_i} = \frac{n(y_i, x_i) + \theta_0}{n(y_i) + |W_{y_i}|\theta_0}. \quad (4)$$

Here, $n(y_i)$ is the number of occurrences of the tag $y_i$ in $\mathbf{y}_{-i}$, $n(y_i, x_i)$ is the number of occurrences of the tag-word pair $(y_i, x_i)$ in $(\mathbf{y}_{-i}, \mathbf{x}_{-i})$, and $W_{y_i}$ is the set of word types in the vocabulary $W$ that can take tag $y_i$. The integral is tractable due to Dirichlet-multinomial conjugacy, and an identical marginalization was applied in the monolingual Bayesian HMM of Goldwater and Griffiths (2007).

For unaligned tags, it is also possible to exactly marginalize the parameter $\phi$ governing transitions. For the transition from $i - 1$ to $i$,

$$P(y_i | y_{i-1}, \mathbf{y}_{-i}, \phi_0) = \int_{\phi_{y_{i-1}}} \phi_{y_{i-1}}(y_i) P(\phi_{y_i} | \mathbf{y}_{-i}, \phi_0) d\phi_{y_{i-1}} = \frac{n(y_{i-1}, y_i) + \phi_0}{n(y_{i-1}) + |T|\phi_0}. \quad (5)$$

The factors here are similar to the emission probability: $n(y_i)$ is the number of occurrences of the tag $y_i$ in $\mathbf{y}_{-i}$, $n(y_{i-1}, y_i)$ is the number of occurrences of the tag sequence $(y_{i-1}, y_i)$, and $T$ is the tagset. The probability for the transition from $i$ to $i + 1$ is analogous.

**Jointly Sampling Aligned Tags** The situation for tags of aligned words is more complex. We sample these tags jointly, considering all $|T \times T'|$ possibilities. We begin by decomposing the probability into three factors:

$$P(y_i, y'_j | \mathbf{y}_{-i}, \mathbf{y}'_{-j}, \mathbf{x}, \mathbf{x}', \mathbf{a}, \theta_0, \theta'_0, \phi, \phi', \omega) \propto P(x_i | \mathbf{y}, \mathbf{x}_{-i}, \theta_0) P(x'_j | \mathbf{y}', \mathbf{x}'_{-j}, \theta'_0) P(y_i, y'_j | \mathbf{y}_{-i}, \mathbf{y}'_{-j}, \mathbf{a}, \phi, \phi', \omega).$$

The first two factors are emissions, and are handled identically to the unaligned case (Formula 4). The expansion of the final, joint factor depends on the alignment of the succeeding tags. If neither of the successors (in either language) are aligned, we have a product of the bilingual coupling probability and four transition probabilities:

$$P(y_i, y'_j | \mathbf{y}_{-i}, \mathbf{y}'_{-j}, \phi, \phi', \omega) \propto \omega(y_i, y'_j) \phi_{y_{i-1}}(y_i) \phi_{y_i}(y_{i+1}) \phi'_{y'_{j-1}}(y'_j) \phi'_{y'_j}(y'_{j+1}).$$

Whenever one or more of the succeeding words is aligned, the sampling formulas must account for the effect of the sampled tag on the joint probability of the succeeding tags, which is no longer a simple multinomial transition probability. We give the formula for one such case—when we are sampling a joint tag pair $(y_i, y'_j)$, whose succeeding words $(x_{i+1}, x'_{j+1})$ are also aligned to one another:

$$P(y_i, y_j | \mathbf{y}_{-i}, \mathbf{y}'_{-j}, \mathbf{a}, \phi, \phi', \omega) \propto \omega(y_i, y'_j) \phi_{y_{i-1}}(y_i) \, \phi'_{y'_{j-1}}(y'_j) \left[ \frac{\phi_{y_i}(y_{i+1}) \, \phi'_{y'_j}(y'_{j+1})}{\sum_{t,t'} \phi_{y_i}(t) \, \phi'_{y'_j}(t') \, \omega(t, t')} \right]. \quad (6)$$

Intuitively, if $\omega$ puts all of its probability mass on a single assignment $y_{i+1} = t, y'_{j+1} = t'$, then the transitions from $i$ to $i + 1$ and $j$ to $j + 1$ are irrelevant, and the final factor goes to one. Conversely, if $\omega$ is indifferent and assigns equal probability to all pairs $\langle t, t' \rangle$, then the final factor becomes proportional to $\phi_{y_i}(y_{i+1}) \phi'_{y'_j}(y'_{j+1})$, which is the same as if $x_{i+1}$ and $x_{i+1}$ were not aligned. In general, as the entropy of $\omega$ increases, the transition to the succeeding nodes exerts a greater influence on $y_i$ and $y'_j$. Similar equations can be derived for cases where the succeeding tags are not aligned to each other, but one of them is aligned to another tag, e.g., $x_{i+1}$ is aligned to $x'_{j+2}$.

As before, we would like to marginalize the parameters $\phi$, $\phi'$, and $\omega$. Because these parameters interact as a product-of-experts model, these marginalizations are approximations. The form of the marginalizations for $\phi$ and $\phi'$ are identical to Formula 5. For the coupling distribution,

$$P_\omega(y_i, y'_j | \mathbf{y}_{-i}, \mathbf{y}'_{-j}, \omega_0) \approx \frac{n(y_i, y'_j) + \omega_0}{N(\mathbf{a}) + |T \times T'|\omega_0}, \quad (7)$$

where $n(y_i, y'_j)$ is the number of times tags $y_i$ and $y'_j$ were aligned, excluding $i$ and $j$, and $N(\mathbf{a})$ is the total number of alignments. As above, the prior $\omega_0$ appears as a smoothing factor; in the denominator it is multiplied by the dimensionality of $\omega$, which is the size of the cross-product of the two tagsets. Intuitively, this approximation would be exactly correct if each aligned tag had been generated twice — once by the transition parameter and once by the coupling parameter — instead of a single time by the product of experts.

The alternative to approximately marginalizing these parameters would be to sample them using a Metropolis-Hastings scheme as in the work by Snyder, Naseem, Eisenstein, and Barzilay (2008). The use of approximate marginalizations represents a bias-variance tradeoff, where the decreased sampling variance justifies the bias introduced by the approximations, for practical numbers of samples.

### 3.2 Multilingual Unsupervised Tagging: A Latent Variable Model

The model described in the previous section is designed for bilingual aligned data; as we will see in Section 5, it exploits such data very effectively. However, many resources contain more than two languages: for example, Europarl contains eleven, and the Multext-East corpus contains eight. This raises the question of how best to exploit all available resources when multi-aligned data is available.

One possibility would be to train separate bilingual models and then combine their output at test time, either by voting or some other heuristic. However, we believe that cross-lingual information reduces ambiguity at training time, so it would be preferable to learn from multiple languages jointly during training. Indeed, the results in Section 5 demonstrate that joint training outperforms such a voting scheme.

Another alternative would be to try to extend the bilingual model developed in the previous section. While such an extension is possible in principle, the merged node model does not scale well in the case of multi-aligned data across more than two languages. Recall that we use merged nodes to represent both tags for aligned words; the state space of such nodes grows as $|T|^L$, exponential in the number of languages $L$. Similarly, the coupling parameter $\omega$ has the same dimension, so that the counts required for estimation become too sparse as the number of languages increases. Moreover, the bi-tag model required removing crossing edges in the word-alignment, so as to avoid cycles. This is unproblematic for pairs of aligned sentences, usually requiring the removal of less than 5% of all edges (see Table 16 in Appendix A). However, as the number of languages grows, an increasing number of alignments will have to be discarded.

Instead, we propose a new architecture specifically designed for the multilingual setting. As before, we maintain HMM substructures for each language, so that the learned parameters can easily be applied to monolingual data. However, rather than merging tag nodes for aligned words, we introduce a layer of *superlingual tags*. The role of these latent nodes is to capture cross-lingual patterns. Essentially they perform a non-parametric clustering over sets of aligned tags, encouraging multilingual patterns that occur elsewhere in the corpus.

More concretely, for every set of aligned words, we add a superlingual tag with outgoing edges to the relevant part-of-speech nodes. An example configuration is shown in Figure 1c. The super-lingual tags are each generated independently, and they influence the selection of the part-of-speech tags to which they are connected. As before, we use a product-of-experts model to combine these cross-lingual cues with the standard HMM transition model.

This setup scales well. Crossing and many-to-many alignments may be used without creating cycles, as all cross-lingual information emanates from the hidden superlingual tags. Furthermore, the size of the model and its parameter space scale linearly with the number of languages. We now describe the role of the superlingual tags in more detail.

### 3.2.1 PROPAGATING CROSS-LINGUAL PATTERNS WITH SUPERLINGUAL TAGS

Each superlingual tag specifies a set of distributions — one for each language's part-of-speech tagset. In order to learn repeated cross-lingual patterns, we need to constrain the number of values that the superlingual tags can take and thus the number of distributions they provide. For example, we might allow the superlingual tags to take on integer values from $1$ to $K$, with each integer value indexing a separate set of tag distributions. Each set of distributions should correspond to a discovered cross-lingual pattern in the data. For example, one set of distributions might favor nouns in each language and another might favor verbs, though heterogenous distributions (e.g., favoring determiners in one language and prepositions in others) are also possible.

Rather than fixing the number of superlingual tag values to an arbitrary size $K$, we leave it unbounded, using a non-parametric Bayesian model. To encourage the desired multilingual clustering behavior, we use a Dirichlet process prior (Ferguson, 1973). Under this prior, high posterior probability is obtained only when a small number of values are used repeatedly. The actual number of sampled values will thus be dictated by the data.

We draw an infinite sequence of distribution sets $\Psi_1, \Psi_2, \ldots$ from some base distribution $G_0$. Each $\Psi_i$ is a set of distributions over tags, with one distribution per language, written $\psi_i^{(\ell)}$. To weight these sets of distributions, we draw an infinite sequence of mixture weights $\pi_1, \pi_2, \ldots$ from a stick-breaking process, which defines a distribution over the integers with most probability mass placed on some initial set of values. The pair of sequences $\pi_1, \pi_2, \ldots$ and $\Psi_1, \Psi_2, \ldots$ now define the distribution over superlingual tags and their associated distributions on parts-of-speech. Each superlingual tag $z \in \mathbb{N}$ is drawn with probability $\pi_z$, and is associated with the set of multinomials $\langle \psi_z^{\ell}, \psi_z^{\ell'}, \ldots \rangle$.

As in the merged node model, the distribution over aligned part-of-speech tags is governed by a product of experts. In this case, the incoming edges are from the superlingual tags (if any) and the predecessor tag. We combine these distributions via their normalized product. Assuming tag position $i$ of language $\ell$ is connected to $M$ superlingual tags, the part-of-speech tag $y_i$ is drawn according to,

$$y_i \sim \frac{\phi_{y_{i-1}}(y_i) \prod_{m=1}^{M} \psi_{z_m}^{\ell}(y_i)}{Z}, \tag{8}$$

where $\phi_{y_{i-1}}$ indicates the transition distribution, $z_m$ is the value of the $m^{th}$ connected superlingual tag, and $\psi_{z_m}^{\ell}(y_i)$ indicates the tag distribution for language $\ell$ given by $\Psi_{z_m}$. The normalization $Z$ is obtained by summing this product over all possible values of $y_i$.

This parameterization allows for a relatively simple parameter space. It also leads to a desirable property: for a tag to have high probability, *each* of the incoming distributions must allow it. That is, any expert can "veto" a potential tag by assigning it low probability, generally leading to consensus decisions.

We now formalize this description by giving the stochastic generative process for the observed data (raw parallel text and alignments), according to the multilingual model.

### 3.2.2 LATENT VARIABLE MODEL: GENERATIVE STORY

For $n$ languages, we assume the existence of $n$ tagsets $T^1, \ldots, T^n$ and vocabularies, $W^1, \ldots, W^n$, one for each language. Table 1 summarizes all relevant parameters. For clarity the generative process is described using only bigram transition dependencies, but our experiments use a trigram model, without any bigram interpolations.

1. **Transition and Emission Parameters**. For each language $\ell = 1, ..., n$ and for each tag $t \in T^\ell$, draw a *transition* distribution $\phi_t^\ell$ over tags $T_\ell$ and an *emission* distribution $\theta_t^\ell$ over words $W^\ell$, all from symmetric Dirichlet priors of appropriate dimension.

2. **Superlingual Tag Parameters**. Draw an infinite sequence of sets of distributions over tags $\Psi_1, \Psi_2, \ldots$, where each $\Psi_i$ is a set of $n$ multinomials $\langle \psi_i^1, \psi_i^2, \ldots \psi_i^n \rangle$, one for each of $n$ languages. Each multinomial $\psi_i^\ell$ is a distribution over the tagset $T^\ell$, and is drawn from a symmetric Dirichlet prior; these priors together comprise the base distribution $G_0$, from which each $\Psi_i$ is drawn.

   At the same time, draw an infinite sequence of mixture weights $\pi \sim GEM(\alpha)$, where $GEM(\alpha)$ indicates the stick-breaking distribution (Sethuraman, 1994) with concentration parameter $\alpha = 1$. These parameters define a distribution over superlingual tags, or equivalently over the part-of-speech distributions that they index:

$$z \quad \sim \quad \sum_k^\infty \pi_k \delta_{k=z} \tag{9}$$

$$\Psi \quad \sim \quad \sum_k^\infty \pi_k \delta_{\Psi = \Psi_k} \tag{10}$$

   where $\delta_{\Psi = \Psi_k}$ is defined as one when $\Psi = \Psi_k$ and zero otherwise. From Formula 10, we can say that the set of multinomials $\Psi$ is drawn from a Dirichlet process, conventionally written $DP(\alpha, G_0)$.

3. **Data**. For each multilingual parallel sentence:

   (a) Draw an alignment **a** from multilingual alignment distribution $A_m$. The alignment **a** specifies sets of aligned indices across languages; each such set may consist of indices in any subset of the languages.

   (b) For each set of indices in **a**, draw a superlingual tag value $z$ according to Formula 9.

   (c) For each language $\ell$, for $i = 1, \ldots$ (until end-tag reached):

       i. Draw a part-of-speech tag $y_i \in T^\ell$ according to Formula 8.

       ii. Draw a word $w_i \in W^\ell$ according to the emission distribution $\theta_{y_i}$.

One important difference from the merged node model generative story is that the distribution over multilingual alignments $A_m$ is unconstrained: we can generate crossing and many-to-one alignments as needed. To perform Bayesian inference under this model we again use Gibbs sampling, marginalizing parameters whenever possible.

### 3.2.3 LATENT VARIABLE MODEL: INFERENCE

As in section 3.1.2, we employ a sampling-based inference procedure. Again, standard closed forms are used to analytically marginalize the emission parameters $\boldsymbol{\theta}$, and approximate marginalizations are applied to transition parameters $\boldsymbol{\phi}$, and superlingual tag distributions $\psi_i^\ell$; similar techniques are used to marginalize the superlingual tag mixture weights $\boldsymbol{\pi}$. As before, these approximations would be exact if each of the parameters in the numerator of Formula 8 were solely responsible for other sampled tags.

We still must sample the part-of-speech tags $\mathbf{y}$ and superlingual tags $\mathbf{z}$. The remainder of the section describes the sampling equations for these variables.

**Sampling Part-of-speech Tags** To sample the part-of-speech tag for language $\ell$ at position $i$ we draw from:

$$P(y_i^\ell|\mathbf{y}_{-(\ell,i)}, \mathbf{x}, \mathbf{a}, \mathbf{z}) \propto P(x_i^\ell|\mathbf{x}_{-i}^\ell, \mathbf{y}^\ell)P(y_{i+1}^\ell|y_i^\ell, \mathbf{y}_{-(\ell,i)}, \mathbf{a}, \mathbf{z})P(y_i^\ell|\mathbf{y}_{-(\ell,i)}, \mathbf{a}, \mathbf{z}) \tag{11}$$

where $\mathbf{y}_{-(\ell,i)}$ refers to all tags except $y_i^\ell$. The first factor handles the emissions, and the latter two factors are the generative probabilities of *(i)* the current tag given the previous tag and superlingual tags, and *(ii)* the next tag given the current tag and superlingual tags. These two quantities are similar to equation 8, except here we integrate over the transition parameter $\phi_{y_{i-1}}$ and the superlingual tag parameters $\omega_z^\ell$. We end up with a product of integrals, each of which we compute in closed form.

Terms involving the transition distributions $\boldsymbol{\phi}$ and the emission distributions $\boldsymbol{\theta}$ are identical to the bilingual case, as described in Section 3.1.2. The closed form for integrating over the parameter of a superlingual tag with value $z$ is given by:

$$\int \psi_z^\ell(y_i)P(\psi_z^\ell|\psi_0^\ell)d\psi_z^\ell = \frac{n(z, y_i, \ell) + \psi_0^\ell}{n(z, \ell) + T^\ell\psi_0^\ell}$$

where $n(z, y_i, \ell)$ is the number of times that tag $y_i$ is observed together with superlingual tag $z$ in language $\ell$, $n(z, \ell)$ is the total number of times that superlingual tag $z$ appears with an edge into language $\ell$, and $\psi_0^\ell$ is a symmetric Dirichlet prior over tags for language $\ell$.

**Sampling Superlingual Tags** For each set of aligned words in the observed alignment $\mathbf{a}$ we need to sample a superlingual tag $z$. Recall that $z$ is an index into an infinite sequence

$$\langle \psi_1^{\ell_1}, \ldots, \psi_1^{\ell_n} \rangle, \langle \psi_2^{\ell_1}, \ldots, \psi_2^{\ell_n} \rangle, \ldots,$$

where each $\psi_z^\ell$ is a distribution over the tagset $T^\ell$. The generative distribution over $z$ is given by Formula 9. In our sampling scheme, however, we integrate over all possible settings of the mixture weights $\boldsymbol{\pi}$ using the standard Chinese Restaurant Process closed form (Escobar & West, 1995):

$$P(z_i|\mathbf{z}_{-i}, \mathbf{y}) \propto \prod_\ell P(y_i^\ell|z_i, \mathbf{z}_{-i}, \mathbf{y}_{-(\ell,i)}) \cdot \begin{cases} \frac{1}{k+\alpha}n(z_i) & \text{if } z_i \in \mathbf{z}_{-i} \\ \frac{\alpha}{k+\alpha} & \text{otherwise} \end{cases} \tag{12}$$

The first group of factors is the product of closed form probabilities for all tags connected to the superlingual tag, conditioned on $z_i$. Each of these factors is calculated in the same manner as equation 11 above. The final factor is the standard Chinese Restaurant Process closed form for posterior sampling from a Dirichlet process prior. In this factor, $k$ is the total number of sampled superlingual tags, $n(z_i)$ is the total number of times the value $z_i$ occurs in the sampled superlingual tags, and $\alpha$ is the Dirichlet process concentration parameter (see Step 2 in Section 3.2.2).

### 3.3 Implementation

This section describes implementation details that are necessary to reproduce our experiments. We present details for the merged node and latent variable models, as well as our monolingual baseline.

#### 3.3.1 INITIALIZATION

An initialization phase is required to generate initial settings for the word tags and hyperparameters, and for the superlingual tags in the latent variable model. The initialization is as follows:

- **Monolingual Model**

  - **Tags:** Random, with uniform probability among tag dictionary entries for the emitted word.
  - **Hyperparameters** $\theta_0$, $\phi_0$**:** Initialized to 1.0

- **Merged Node Model**

  - **Tags**: Random, with uniform probability among tag dictionary entries for the emitted word. For joined tag nodes, each slot is selected from the tag dictionary of the emitted word in the appropriate language.
  - **Hyperparameters** $\theta_0$, $\phi_0$, $\omega_0$**:** Initialized to 1.0

- **Latent Variable Model**

  - **Tags:** Set to the final estimate from the monolingual model.
  - **Superlingual Tags:** Initially a set of 14 superlingual tag values is assumed — each value corresponds to one part-of-speech tag. Each alignment is assigned one of these 14 values based on the most common initial part-of-speech tag of the words in the alignment.
  - **Hyperparameters** $\theta_0^\ell$, $\phi_0^\ell$**:** Initialized to 1.0
  - **Base Distribution** $G_0^\ell$**:** Set to a symmetric Dirichlet distribution with parameter value fixed to 1.0
  - **Concentration Parameter** $\alpha$: Set to 1.0 and remains fixed throughout.

#### 3.3.2 HYPERPARAMETER ESTIMATION

Both models have symmetric Dirichlet priors $\theta_0$ and $\phi_0$, for the emission and transition distributions respectively. The merged node model also has symmetric Dirichlet prior $\omega_0$ on the coupling parameter. We re-estimate these priors during inference, based on non-informative hyperpriors.

Hyperparameter re-estimation applies the Metropolis-Hastings algorithm after each full epoch of sampling the tags. In addition, we run an initial 200 iterations to speed convergence. Metropolis-Hastings is a sampling technique that draws a new value $u$ from a proposal distribution, and makes a stochastic decision about whether to accept the new sample (Gelman et al., 2004). This decision is based on the proposal distribution and on the joint probability of $u$ with the observed and sampled variables $\mathbf{x}^\ell$ and $\mathbf{y}^\ell$.

We assume an improper prior $P(u)$ that assigns uniform probability mass over the positive reals, and use a Gaussian proposal distribution with the mean set to the previous value of the parameter and variance set to one-tenth of the mean.[5] For non-pathological proposal distributions, the Metropolis-Hastings algorithm is guaranteed to converge in the limit to a stationary Markov chain centered on the desired joint distribution. We observe an acceptance rate of approximately 1/6, which is in line with standard recommendations for rapid convergence (Gelman et al., 2004).

### 3.3.3 FINAL PARAMETER ESTIMATES

The ultimate goal of training is to learn models that can be applied to unaligned monolingual data. Thus, we need to construct estimates for the transition and emission parameters $\phi$ and $\theta$. Our sampling procedure focuses on the tags $\mathbf{y}$. We construct maximum *a posteriori* estimates $\hat{y}$, indicating the most likely tag sequences for the aligned training corpus. The predicted tags $\hat{y}$ are then combined with priors $\phi_0$ and $\theta_0$ to construct maximum *a posteriori* estimates of the transition and emission parameters. These learned parameters are then applied to the monolingual test data to find the highest probability tag sequences using the Viterbi algorithm.

For the monolingual and merged node models, we perform 200 iterations of sampling, and select the modal tag settings in each slot. Further sampling was not found to produce different results. For the latent variable model, we perform 1000 iterations of sampling, and select the modal tag values from the last 100 samples.

## 4. Experimental Setup

We perform a series of empirical evaluations to quantify the contribution of bilingual and multilingual information for unsupervised part-of-speech tagging. Our first evaluation follows the standard procedures established for unsupervised part-of-speech tagging: given a tag dictionary (i.e., a set of possible tags for each word type), the model selects the appropriate tag for each token occurring in a text. We also evaluate tagger performance when the available dictionaries are incomplete (Smith & Eisner, 2005; Goldwater & Griffiths, 2007). In all scenarios, the model is trained using only untagged text.

In this section, we first describe the parallel data and part-of-speech annotations used for system evaluation. Next we describe a monolingual baseline and the inference procedure used for testing.

### 4.1 Data

As a source of parallel data, we use Orwell's novel "Nineteen Eighty Four" in the original English as well as its translation to seven languages — Bulgarian, Czech, Estonian, Hungarian, Slovene, Serbian and Romanian.[6] Each translation was produced by a different translator and published in print separately by different publishers.

This dataset has representatives from four language families — Slavic, Romance, Ugric and Germanic. This data is distributed as part of the publicly available Multext-East corpus, Version 3 (Erjavec, 2004). The corpus provides detailed morphological annotation at the token level, including part-of-speech tags. In addition, a lexicon for each language is provided.

---

5. This proposal is identical to the parameter re-estimation applied for emission and transition priors by Goldwater and Griffiths (2007).

6. In our initial publication (Snyder et al., 2008), we used a subset of this data, only including sentences that have one-to-one alignments between all four languages considered in that paper. The current set-up makes use of all the sentences available in the corpus.

| | Sentences | Words | Percentage Aligned | | | | | | | |
|---|---|---|---|---|---|---|---|---|---|---|
| | | | BG | CS | EN | ET | HU | RO | SL | SR |
| Bulgarian (BG) | 6681 | 101175 | - | 41.7 | 50.5 | 33.5 | 31.3 | 41.5 | 45.4 | 45.9 |
| Czech (CS) | 6750 | 102834 | 41.0 | - | 41.9 | 39.1 | 30.7 | 31.7 | 56.2 | 48.4 |
| English (EN) | 6736 | 118426 | 43.2 | 36.4 | - | 34.4 | 32.9 | 42.5 | 44.6 | 40.9 |
| Estonian (ET) | 6477 | 94900 | 35.7 | 42.4 | 42.9 | - | 33.8 | 29.2 | 44.8 | 39.7 |
| Hungarian (HU) | 6767 | 98428 | 32.2 | 32.0 | 39.6 | 32.6 | - | 26.9 | 34.6 | 30.3 |
| Romanian (RO) | 6519 | 118330 | 35.5 | 27.5 | 42.5 | 23.4 | 22.4 | - | 30.8 | 32.1 |
| Slovene (SL) | 6688 | 116908 | 39.3 | 49.4 | 45.2 | 36.4 | 29.1 | 31.2 | - | 51.2 |
| Serbian (SR) | 6676 | 112131 | 41.4 | 44.4 | 43.2 | 33.6 | 26.6 | 33.9 | 53.4 | - |

Table 2: Percentage of the words in the row language that have alignments when paired with the column language.

The corpus consists of 118,426 English words in 6,736 sentences (see Table 3). Of these sentences, the first 75% are used for training, taking advantage of the multilingual alignments. The remaining 25% are used for evaluation. In the evaluation, only monolingual information is made available to the model, to simulate performance on non-parallel data.

## 4.2 Alignments

In our experiments we use sentence-level alignments provided in the Multext-East corpus. Word-level alignments are computed for each language pair using GIZA++ (Och & Ney, 2003). The procedures for handling these alignments are different for the merged node and latent variable models.

### 4.2.1 Merged Node Model

We obtain 28 parallel bilingual corpora by considering all pairings of the eight languages. To generate one-to-one alignments at the word level, we intersect the one-to-many alignments going in each direction. This process results in alignment of about half the tokens in each bilingual parallel corpus. We further automatically remove crossing alignment edges, as these would induce cycles in the graphical model. We employ a simple heuristic: crossing alignment edges are removed based on the order in which they appear from left to right; this step eliminates on average 3.62% of the edges. Table 2 shows the number of aligned words for each language pair after removing crossing edges. More detailed statistics about the total number of alignments are provided in Appendix A.

### 4.2.2 Latent Variable Model

As in the previous setting, we run GIZA++ on all 28 pairings of the 8 languages, taking the intersection of alignments in each direction. Since we want each latent superlingual variable to span as many languages as possible, we aggregate pairwise lexical alignments into larger sets of densely aligned words and assign a single latent superlingual variable to each such set. Specifically, for each word token, we consider the set of the word itself and all word tokens to which it is aligned. If pairwise alignments occur between 2/3 of all token pairs in this set, then it is considered densely connected and is admitted as an alignment set. Otherwise, increasingly smaller subsets are considered until one that is densely connected is found. This procedure is repeated for all word tokens in the corpus that have at least one alignment. Finally, the alignment sets are pruned by removing those which are subsets of larger alignment sets. Each of the remaining sets is considered the site of a latent superlingual variable.

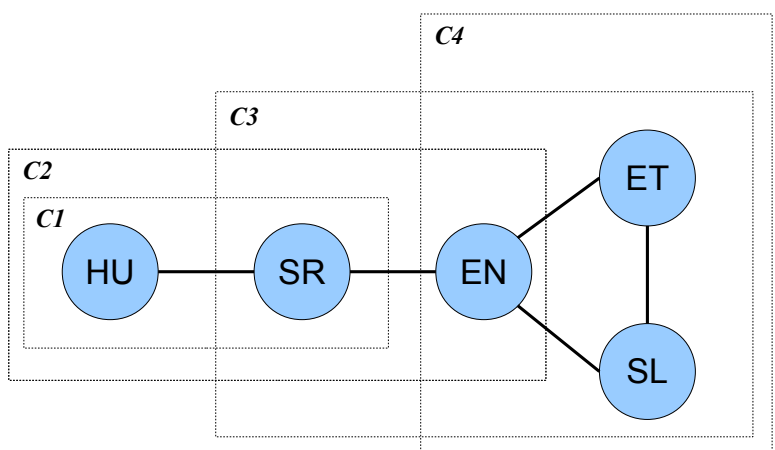

Figure 2: An example of a multilingual alignment configuration. Nodes correspond to words tokens, and are labeled by their language. Edges indicate pairwise alignments produced by GIZA++. Boxes indicate alignment sets, though the set **C1** is subsumed by **C2** and eventually discarded, as described in the text.

This process can be illustrated by an example. The sentence "I know you, the eyes seemed to say, I see through you," appears in the original English version of the corpus. The English word token *seemed* is aligned to word tokens in Serbian (*činilo*), Estonian (*näis*), and Slovenian (*zdelo*). The Estonian and Slovenian tokens are aligned to each other. Finally, the Serbian token is aligned to a Hungarian word token (*mintha*), which is itself not aligned to any other tokens. This configuration is shown in Figure 2, with the nodes labeled by the two-letter language abbreviations.

We now construct alignment sets for these words.

- For the Hungarian word, there is only one other aligned word, in Serbian, so the alignment set consists only of this pair (**C1** in the figure).

- The Serbian word has aligned partners in both Hungarian and English; overall this set has two pairwise alignments out of a possible three, as the English and Hungarian words are not aligned. Still, since 2/3 of the possible alignments are present, an alignment set (**C2**) is formed. **C1** is subsumed by **C2**, so it is eliminated.

- The English word is aligned to tokens in Serbian, Estonian, and Slovenian; four of six possible links are present, so an alignment set (**C3**) is formed. Note that if the Estonian and Slovenian words were not aligned to each other then we would have only three of six links, so the set would not be densely connected by our definition; we would then remove a member of the alignment set.

- The Estonian token is aligned to words in Slovenian and English; all three pairwise alignments are present, so an alignment set (*C4*) is formed. An identical alignment set is formed by starting with the Slovenian word, but only one superlingual tag is created.

Thus, for these five word tokens, a total of three overlapping alignment sets are created. Over the entire corpus, this process results in 284,581 alignment sets, covering 76% of all word tokens. Of these tokens, 61% occur in exactly one alignment set, 29% occur in exactly two alignment sets, and the remaining 10% occur in three or more alignment sets. Of all alignment sets, 32% include words in just two languages, 26% include words in exactly three languages, and the remaining 42% include words in four or more languages. The sets remain fixed during sampling and are treated by the model as observed data.

| | Number of Tokens | Tags per token when lexicon contains ... | | | | Trigram Entropy |
|---|---|---|---|---|---|---|
| | | all words | count $> 5$ | count $> 10$ | top 100 words | |
| Bulgarian (BG) | 101175 | 1.39 | 4.61 | 5.48 | 7.33 | 1.63 |
| Czech (CS) | 102834 | 1.35 | 5.27 | 6.37 | 8.24 | 1.64 |
| English (EN) | 118426 | 1.49 | 3.11 | 3.81 | 6.21 | 1.51 |
| Estonian (ET) | 94900 | 1.36 | 4.91 | 5.82 | 7.34 | 1.61 |
| Hungarian (HU) | 98428 | 1.29 | 5.42 | 6.41 | 7.85 | 1.62 |
| Romanian (RO) | 118330 | 1.55 | 4.49 | 5.53 | 8.54 | 1.73 |
| Slovene (SL) | 116908 | 1.33 | 4.59 | 5.49 | 7.23 | 1.64 |
| Serbian (SR) | 112131 | 1.38 | 4.76 | 5.73 | 7.61 | 1.73 |

Table 3: Corpus size and tag/token ratio for each language in the set. The last column shows the trigram entropy for each language based on the annotations provided with the corpus.

## 4.3 Tagset

The Multext-East corpus is manually annotated with detailed morphosyntactic information. In our experiments, we focus on the main syntactic category encoded as the first letter of the provided labels. The annotation distinguishes between 14 parts-of-speech, of which 11 are common for all languages in our experiments. Appendix B lists the tag repository for each of the eight languages.

In our first experiment, we assume that a complete tag lexicon is available, so that the set of possible parts-of-speech for each word is known in advance. We use the tag dictionaries provided in the Multext-East corpus. The average number of possible tags per token is 1.39. We also experimented with incomplete tag dictionaries, where entries are only available for words appearing more than five or ten times in the corpus. For other words, the entire tagset of 14 tags is considered. In these two scenarios, the average per-token tag ambiguity is 4.65 and 5.58, respectively. Finally we also considered the case when lexicon entries are available for only the 100 most frequent words. In this case the average tags per token ambiguity is 7.54. Table 3 shows the specific tag/token ratio for each language for all scenarios.

In the Multext-East corpus, punctuation marks are not annotated with part-of-speech tags. We expand the tag repository by defining a separate tag for all punctuation marks. This allows the model to make use of any transition or coupling patterns involving punctuation marks. However, we do not consider punctuation tokens when computing model accuracy.

### 4.4 Monolingual Comparisons

As our monolingual baseline we use the unsupervised Bayesian hidden Markov model (HMM) of Goldwater and Griffiths (2007). This model, which they call BHMM1, modifies the standard HMM by adding priors and by performing Bayesian inference. Its performance is on par with state-of-the-art unsupervised models. The Bayesian HMM is a particularly informative baseline because our model reduces to this baseline when there are no alignments in the data. This implies that any performance gain over the baseline can only be attributed to the multilingual aspect of our model. We used our own implementation after verifying that its performance on the Penn Treebank corpus was identical to that reported by Goldwater and Griffiths.

To provide an additional point of comparison, we use a supervised hidden Markov model trained using the annotated corpus. We apply the standard maximum-likelihood estimation and perform inference using Viterbi decoding with pseudo-count smoothing for unknown words (Rabiner, 1989). In Appendix C we also report supervised results using the "Stanford Tagger", version 1.6[7]. Although the results are slightly lower than our own supervised HMM implementation, we note that this system is not directly comparable to our set-up, as it does not allow the use of a tag dictionary to constrain part-of-speech selections.

### 4.5 Test Set Inference

We use the same procedure to apply all the models (the monolingual model, the bilingual merged node model, and the latent variable model) to test data. After training, trigram transition and word emission probabilities are computed, using the counts of tags assigned in the final training iteration. Similarly, the final sampled values of the hyperparameters are selected as smoothing parameters. We then apply Viterbi decoding to identify the highest probability tag sequences for each monolingual test set. We report results for multilingual and monolingual experiments averaged over five runs and for bilingual experiments averaged over three runs. The average standard-deviation of accuracy over multiple runs is less than 0.25 except when the lexicon is limited to the 100 most frequent words. In that case the standard deviation is 1.11 for monolingual model, 0.85 for merged node model and 1.40 for latent variable model.

## 5. Results

In this section, we first report the performance for the two models on the full and reduced lexicon cases. Next, we report results for a semi-supervised experiment, where a subset of the languages have annotated text at training time. Finally, we investigate the sensitivity of both models to hyper-parameter values and provide run time statistics for the latent variable model for increasing numbers of languages.

---

7. http://nlp.stanford.edu/software/tagger.shtml

|  | **Avg** | BG | CS | EN | ET | HU | RO | SL | SR |
|---|---|---|---|---|---|---|---|---|---|
| 1. Random | 83.3 | 82.5 | 86.9 | 80.7 | 84.0 | 85.7 | 78.2 | 84.5 | 83.5 |
| 2. Monolingual | 91.2 | 88.7 | 93.9 | 95.8 | 92.7 | 95.3 | 91.1 | 87.4 | 84.5 |
| 3. MERGEDNODE: *average* | **93.2** | 91.3 | 96.9 | 95.9 | 93.3 | 96.7 | 91.9 | 89.3 | 90.2 |
| 4. LATENTVARIABLE | **95.0** | 92.6 | 98.2 | 95.0 | 94.6 | 96.7 | 95.1 | 95.8 | 92.3 |
| 5. Supervised | 97.3 | 96.8 | 98.6 | 97.2 | 97.0 | 97.8 | 97.7 | 97.0 | 96.6 |
| 6. MERGEDNODE: *voting* | 93.0 | 91.6 | 97.4 | 96.1 | 94.3 | 96.8 | 91.6 | 87.9 | 88.2 |
| 7. MERGEDNODE: *best pair* | 95.4 | 94.7 | 97.8 | 96.1 | 94.2 | 96.9 | 94.1 | 94.8 | 94.5 |

Table 4: Tagging accuracy with complete tag dictionaries. The first column reports average results across all languages (see Table 3 for language name abbreviations). The latent variable model is trained using all eight languages, whereas the merged node models are trained on language pairs. In the latter case, results are given by averaging over all pairings (line 3), by having all bilingual models vote on each tag prediction (line 6), and by having an oracle select the best pairing for each target language (line 7). All differences between LATENT-VARIABLE, MERGEDNODE: *voting*, and Monolingual (lines 2, 4, and 6) are statistically significant at $p < 0.05$ according to a sign test.

## 5.1 Full Lexicon Experiments

Our experiments show that both the merged node and latent variable models substantially improve tagging accuracy. Since the merged node model is restricted to pairs of languages, we provide average results over all possible pairings. In addition, we also consider two methods for combining predictions from multiple bilingual pairings: one using a voting scheme and the other employing an oracle to select the best pairings (see below for additional details).

As shown in Line 4 of Table 4, the merged node model achieves, on average, 93.2% accuracy, a two percentage point improvement over the monolingual baseline.[8] The latent variable model — trained once on all eight languages — achieves 95% accuracy, nearly two percentage points higher than the bilingual merged node model. These two results correspond to error reductions of 23% and 43% respectively, and reduce the gap between unsupervised and supervised performance by over 30% and 60%.

As mentioned above, we also employ a voting scheme to combine information from multiple languages using the merged node model. Under this scheme, we train bilingual merged node models for each language pair. Then, when making tag predictions for a particular language — e.g., Romanian — we consider the preferences of each bilingual model trained with Romanian and a second language. The tag preferred by a plurality of models is selected. The results for this method are shown in line 6 of Table 4, and do not differ significantly from the average bilingual performance. Thus, this simple method of combining information from multiple language does not measure up to the joint multilingual model performance.

---

8. The accuracy of the monolingual English tagger is relatively high compared to the 87% reported by Goldwater and Griffiths (2007) on the WSJ corpus. We attribute this discrepancy to the differences in tag inventory used in our data-set. For example, when *Particles* and *Prepositions* are merged in the WSJ corpus (as they happen to be in our tag inventory and corpus), the performance of Goldwater's model on WSJ is similar to what we report here.

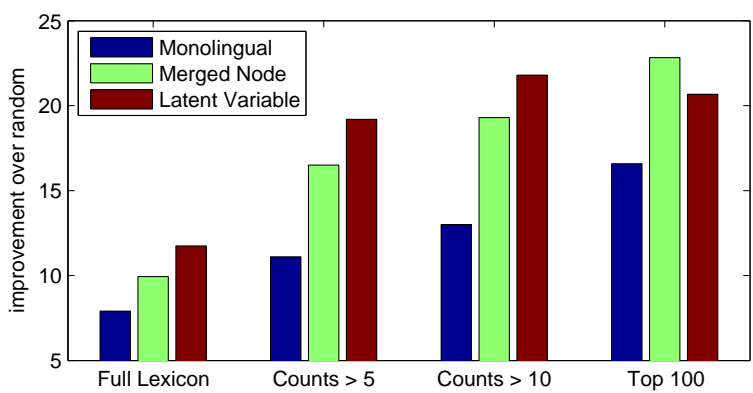

Figure 3: Summary of model performance in full and reduced lexicon conditions. Improvement over the random baseline is indicated for the monolingual baseline, the merged node model (average performance over all possible bilingual pairings), and the latent variable model (trained on all eight languages). "Counts $> x$" indicates that only words with counts greater than $x$ were kept in the lexicon; "Top 100" keeps only the 100 most common words.

We use the sign test to assess whether there are statistically significant differences in the accuracy of the tag predictions made by the monolingual baseline (line 2 of Table 4), the latent variable model (line 4), and the voting-based merged node model (line 6). All differences in these rows are found to be statistically significant at $p < 0.05$. Note that we cannot use the sign test to compare the average performance of the bilingual model (line 3), since this result is an aggregate over accuracies for every language pair.

## 5.2 Reduced Lexicon Experiments

In realistic application scenarios, we may not have a tag dictionary with coverage across the entire lexicon. We consider three reduced lexicons: removing all words with counts of five or less; removing all words with counts of ten or less; and keeping only the top 100 most frequent words. Words that are removed from the lexicon can take any tag, increasing the overall difficulty of the task. These results are shown in Table 5 and graphically summarized in Figure 3. In all cases, the monolingual model is less robust to reduction in lexicon coverage than the multilingual models. In the case of the 100 word lexicon, the latent variable model achieves accuracy of 57.9%, compared to 53.8% for the monolingual baseline. The merged node model, on the other hand, achieves a slightly higher average performance of 59.5%. In the two other scenarios, the latent variable model trained on all eight languages outperforms the bilingual merged node model, even when an oracle selects the best bilingual pairing for each target language. For example, using the lexicon with words that appear greater than five times, the monolingual baseline achieves 74.7% accuracy, the merged node model using the best possible pairings achieves 81.7% accuracy, and the full latent variable model achieves an accuracy of 82.8%.

| | | **Avg** | BG | CS | EN | ET | HU | RO | SL | SR |
|---|---|---|---|---|---|---|---|---|---|---|
| **Counts > 5** | Random | 63.6 | 62.9 | 62 | 71.8 | 61.6 | 61.3 | 62.8 | 64.8 | 61.8 |
| | Monolingual | 74.8 | 73.5 | 72.2 | 87.3 | 72.5 | 73.5 | 77.1 | 75.7 | 66.3 |
| | MERGEDNODE: *average* | 80.1 | 80.2 | 79 | 90.4 | 76.5 | 77.3 | 82.7 | 78.7 | 75.9 |
| | LATENTVARIABLE | **82.8** | 81.3 | **83.0** | 88.1 | **80.6** | **80.8** | **86.1** | **83.6** | 78.8 |
| | MERGEDNODE: *voting* | 80.4 | 80.4 | 78.5 | 90.7 | 76.4 | 76.8 | 84.0 | 79.7 | 76.4 |
| | MERGEDNODE: *best pair* | 81.7 | **82.7** | 79.7 | **90.7** | 77.5 | 78 | 84.4 | 80.9 | **79.4** |
| **Counts > 10** | Random | 57.9 | 57.5 | 54.7 | 68.3 | 56 | 55.1 | 57.2 | 59.2 | 55.5 |
| | Monolingual | 70.9 | 71.9 | 66.7 | 84.4 | 68.3 | 69.0 | 73.0 | 70.4 | 63.7 |
| | MERGEDNODE: *average* | 77.2 | 77.8 | 75.3 | 88.8 | 72.9 | 73.8 | 80.5 | 76.1 | 72.4 |
| | LATENTVARIABLE | **79.7** | 78.8$^\dagger$ | **79.4** | 86.1 | **77.9** | **76.4** | **83.1** | **80.0** | 75.9 |
| | MERGEDNODE: *voting* | 77.5 | 78.4$^\dagger$ | 75.3 | 89.2 | 73.1 | 73.3 | 81.7 | 76.1 | 73.1 |
| | MERGEDNODE: *best pair* | 79.0 | **80.2** | 76.7 | **89.4** | 74.9 | 75.2 | 82.1 | 77.6 | **76.1** |
| **Top 100** | Random | 37.3 | 36.7 | 32.1 | 48.9 | 36.6 | 36.4 | 33.7 | 39.8 | 33.8 |
| | Monolingual | 53.8 | 60.9$^\ddagger$ | 44.1 | 69.0 | 54.8* | 56.8 | 51.4 | 49.4 | 44.0 |
| | MERGEDNODE: *average* | **59.6** | 60.1 | 52.5 | 73.5 | 59.5 | 59.4 | 61.4 | 56.6 | 53.4 |
| | LATENTVARIABLE | 57.9 | **65.5** | 49.3 | 71.6 | 54.3* | 51.0 | 57.5 | 53.9 | **60.4** |
| | MERGEDNODE: *voting* | 62.4 | 61.5$^\ddagger$ | 55.4 | 74.8 | 62.2 | 60.9 | 64.3 | 62.3 | 57.5 |
| | MERGEDNODE: *best pair* | 63.6 | 64.7 | **55.3** | **77.4** | **61.5** | **60.2** | **69.3** | **63.1** | 56.9 |

Table 5: Tagging accuracy in reduced lexicon conditions. "Counts > $x$" indicates that only words with counts greater than $x$ were kept in the lexicon; "Top 100" keeps only the 100 most common words. The latent variable model is trained using all eight languages, whereas the merged node models are trained on language pairs. In the latter case, results are given by averaging over all pairings, by having all bilingual models vote on each tag prediction, and by having an oracle select the best pairing for each target language. Other than the three pairs of results marked with $\dagger$, $\ddagger$, and $*$, all differences between "monolingual", "LATENTVARIABLE", and "MERGEDNODE: *voting*" are statistically significant at $p < 0.05$ according to a sign test.

Next we consider the performance of the bilingual merged node model when the lexicon is reduced for only one of the two languages. This condition may occur when dealing with two languages with asymmetric resources, in terms of unannotated text. As shown in Table 6, the merged models on average scores 5.7 points higher than the monolingual model when both tag dictionaries are reduced, but 14.3 points higher when the partner language has a full tag dictionary. This suggests that the bilingual models effectively transfer the additional lexical information available for the resource-rich language to the resource-poor language, yielding substantial performance improvements.

Perhaps the most surprising result is that the resource-rich language gains as much on average from pairing with the resource-poor partner language as it would have gained from pairing with a language with a full lexicon. In both cases, an average accuracy of 93.2% is achieved, compared to the 91.1% monolingual baseline.

| | Monolingual | | Bilingual (Merged Node) | | | |
| | Reduced | Full | Both reduced | Reduced language | Unreduced language | Both full |
|---|---|---|---|---|---|---|
| BG | 60.9 | 88.7 | 60.1 | 71.3 | 91.6 | 91.3 |
| CS | 44.1 | 93.9 | 52.5 | 66.7 | 97.1 | 96.9 |
| EN | 69.0 | 95.8 | 73.5 | 82.4 | 95.8 | 95.9 |
| ET | 54.8 | 92.7 | 59.5 | 65.6 | 93.3 | 93.3 |
| HU | 56.8 | 95.3 | 59.4 | 63.0 | 96.7 | 96.7 |
| RO | 51.4 | 91.1 | 61.4 | 69.3 | 91.5 | 91.9 |
| SL | 49.4 | 87.4 | 56.6 | 63.3 | 89.1 | 89.3 |
| SR | 44.0 | 84.5 | 53.4 | 63.6 | 90.3 | 90.2 |
| Avg. | 53.8 | 91.2 | 59.5 | 68.1 | 93.2 | 93.2 |

Table 6: Various scenarios for reducing the tag dictionary to the 100 most frequent terms.

## 5.3 Indirect Supervision

Although the main focus of this paper is unsupervised learning, we also provide some results indicating that multilingual learning can be applied to scenarios with varying amounts of annotated data. These scenarios are in fact quite realistic, as previously trained and highly accurate taggers will usually be available for at least some of the languages in a parallel corpus. We apply our latent variable model to these scenarios by simply treating the tags of annotated data (in any subset of languages) as fixed and observed throughout the sampling procedure. From a strictly probabilistic perspective this is the correct approach. However, we note that, in practice, heuristics and objective functions which place greater emphasis on the supervised portion of the data may yield better results. We do not explore that possibility here.

| | | supervised language(s)... | | | | | | | | | |
| | | BG | CS | EN | ET | HU | RO | SL | SR | All others | None |
| *accuracy for...* | BG | | 69.1 | 68.0 | 65.9 | 60.4 | 67.1 | 73.9 | 69.6 | 76.2 | 65.5 |
| | CS | 50.8 | | 52.2 | 50.2 | 51.2 | 51.0 | 56.6 | 53.1 | 76.6 | 49.3 |
| | EN | 62.6 | 70.5 | | 68.1 | 61.8 | 61.9 | 80.6 | 69.5 | 82.8 | 71.6 |
| | ET | 57.2 | 58.0 | 57.7 | | 56.1 | 56.4 | 59.8 | 57.1 | 72.5 | 54.3 |
| | HU | 50.3 | 50.0 | 53.1 | 51.4 | | 51.1 | 49.8 | 50.0 | 62.3 | 51.0 |
| | RO | 62.8 | 61.6 | 61.3 | 57.8 | 58.5 | | 62.9 | 59.2 | 74.9 | 57.5 |
| | SL | 55.0 | 56.8 | 55.6 | 53.2 | 54.4 | 54.7 | | 56.2 | 77.7 | 53.9 |
| | SR | 64.9 | 65.9 | 64.1 | 63.5 | 61.6 | 63.4 | 69.9 | | 72.5 | 60.4 |
| | Avg | 57.7 | 61.7 | 58.9 | 58.6 | 57.7 | 57.9 | 64.8 | 59.2 | 74.4 | 57.9 |

Table 7: Performance of the latent variable model when some of the eight languages have supervised annotations and the others have only the most frequent 100 words lexicon. The first eight columns report results when only one of the eight languages is supervised. The penultimate column reports results when all but one of the languages are supervised. The final column reports results when *no* supervision is available (repeated from Table 5 for convenience).

Table 7 gives results for two scenarios of indirect supervision: where only one of the eight languages has annotated data, and where all *but* one of the languages has annotated data. In both cases, the unsupervised languages are provided with a 100 word lexicon, and all eight languages are trained together. When only one of the eight languages is supervised, the results vary depending on the choice of supervised language. When one of Bulgarian, Hungarian, or Romanian is supervised, no improvement is seen, on average, for the other seven languages. However, when Slovene is supervised, the improvement seen for the other languages is fairly substantial, with average accuracy rising to 64.8%, from 57.9% for the unsupervised latent variable model and 53.8% for the monolingual baseline. Perhaps unsurprisingly, the results are more impressive when all *but* one of the languages is supervised. In this case, the average accuracy of the lone unsupervised language rises to 74.4%. Taken together, these results indicate that any mixture of supervised resources may be added to the mix in a very simple and straightforward way, often yielding substantial improvements for the other languages.

### 5.4 Hyperarameter Sensitivity and Runtime Statistics

Both models employ hyperparameters for the emission and transition distribution priors ($\theta_0$ and $\phi_0$ respectively) and the merged node model employs an additional hyperparameter for the coupling distribution prior ($\omega_0$). These hyperparameters are all updated throughout the inference procedure. The latent variable model uses two additional hyperparameters that remained fixed: the concentration parameter of the Dirichlet process ($\alpha$) and the parameter of the base distribution for super-lingual tags ($\psi_0$). For the experiments described above we used the initialization values listed in Section 3.3.1. Here we investigate the sensitivity of the models to different initializations of $\theta_0$, $\phi_0$, and $\omega_0$, and to different fixed values of $\alpha$ and $\psi_0$. Tables 8 and 9 show the results obtained for the merged node and latent variable models, respectively, using a full lexicon. We observe that across a wide range of values, both models yield very similar results. In addition, we note that the final sampled hyperparameter values for transition and emission distributions always fall below one, indicating that sparse priors are preferred.

As mentioned in Section 3.2 one of the key theoretical benefits of the latent variable approach is that the size of the model and its parameter space scale linearly with the number of languages. Here we provide empirical confirmation by running the latent variable model on all possible subsets of the eight languages, recording the time elapsed for each run[9]. Figure 4 shows the average running time as the number of languages is increased (averaged over all subsets of each size). We see that the model running time indeed scales linearly as languages are added, and that the per-language running time increases very slowly: when all eight languages are included, the time taken is roughly double that for eight monolingual models run serially. Both of our models scale well with tagset size and the number of examples. The time dependence on the former is cubic, as we use trigram models and employ Viterbi decoding to find optimal sequences at test-time. During the training time, however, the time scales linearly with the tagset size for the latent variable model and quadratically for the merged node model. This is due to the use of Gibbs sampling that isolates the individual sampling decision on tags (for the latent variable model) and tag-pairs (for the merged node model). The dependence on the number of training examples is also linear for the same reason.

---

9. All experiments were single-threaded and run using an Intel Xeon 3.0 GHz processor

| MERGEDNODE: *hyperparameter initializations* | | | | | | |
|---|---|---|---|---|---|---|
| $\phi_0$ | 1.0 | **0.1** | **0.01** | 1.0 | 1.0 | 1.0 | 1.0 |
| $\theta_0$ | 1.0 | 1.0 | 1.0 | **0.1** | **0.01** | 1.0 | 1.0 |
| $\omega_0$ | 1.0 | 1.0 | 1.0 | 1.0 | 1.0 | **0.1** | **0.01** |
| BG | 91.3 | 91.3 | 91.3 | 91.3 | 91.2 | 91.1 | 91.3 |
| CS | 96.9 | 97.0 | 97.0 | 96.9 | 96.8 | 96.5 | 97.1 |
| EN | 95.9 | 95.9 | 95.9 | 95.9 | 95.9 | 95.9 | 95.9 |
| ET | 93.3 | 93.4 | 93.3 | 93.4 | 93.2 | 93.4 | 93.2 |
| HU | 96.7 | 96.7 | 96.7 | 96.7 | 96.7 | 96.7 | 96.8 |
| RO | 91.9 | 91.8 | 91.8 | 91.9 | 91.8 | 91.8 | 91.8 |
| SL | 89.3 | 89.3 | 89.3 | 89.3 | 89.4 | 89.3 | 89.3 |
| SR | 90.2 | 90.2 | 90.2 | 90.2 | 90.2 | 90.2 | 90.2 |
| Avg | 93.2 | 93.2 | 93.2 | 93.2 | 93.2 | 93.1 | 93.2 |

Table 8: Results for different initializations of the hyperparameters of the merged node model. $\phi_0$, $\theta_0$ and $\omega_0$ are the hyperparameters for the transition, emission and coupling multinomials respectively. The results for each language are averaged over all possible pairings with the other languages.

| LATENTVARIABLE: *hyperparameter initializations & settings* | | | | | | | | | |
|---|---|---|---|---|---|---|---|---|---|
| $\alpha$ | 1.0 | **0.1** | **10** | **100** | 1.0 | 1.0 | 1.0 | 1.0 | 1.0 | 1.0 |
| $\psi_0$ | 1.0 | 1.0 | 1.0 | 1.0 | **0.1** | **0.01** | 1.0 | 1.0 | 1.0 | 1.0 |
| $\phi_0$ | 1.0 | 1.0 | 1.0 | 1.0 | 1.0 | 1.0 | **0.1** | **0.01** | 1.0 | 1.0 |
| $\theta_0$ | 1.0 | 1.0 | 1.0 | 1.0 | 1.0 | 1.0 | 1.0 | 1.0 | **0.1** | **0.01** |
| BG | 92.6 | 92.6 | 92.6 | 92.6 | 92.6 | 92.7 | 92.6 | 92.6 | 92.6 | 92.6 |
| CS | 98.2 | 98.1 | 98.2 | 98.2 | 98.1 | 98.1 | 98.2 | 98.1 | 98.2 | 98.1 |
| EN | 95.0 | 95.0 | 94.9 | 94.8 | 95.1 | 95.2 | 95.0 | 94.9 | 94.9 | 95.0 |
| ET | 94.6 | 95.0 | 95.0 | 94.9 | 94.2 | 94.8 | 95.0 | 94.9 | 94.9 | 94.5 |
| HU | 96.7 | 96.7 | 96.7 | 96.7 | 96.7 | 96.6 | 96.7 | 96.7 | 96.7 | 96.7 |
| RO | 95.1 | 95.0 | 95.1 | 95.1 | 95.2 | 95.1 | 95.0 | 94.9 | 95.1 | 95.0 |
| SL | 95.8 | 95.8 | 95.8 | 95.8 | 95.8 | 95.8 | 95.8 | 95.8 | 95.8 | 95.8 |
| SR | 92.3 | 92.3 | 92.3 | 92.3 | 92.4 | 92.4 | 92.3 | 92.3 | 92.3 | 92.3 |
| Avg | 95.0 | 95.1 | 95.1 | 95.0 | 95.0 | 95.1 | 95.1 | 95.0 | 95.1 | 95.0 |

Table 9: Results for different initializations and settings of hyperparameters of the latent variable model. $\phi_0$ and $\theta_0$ are the hyperparameters for the transition and emission multinomials respectively and are updated throughout inference. $\alpha$ and $\psi_0$ are the concentration parameter and base distribution parameter, respectively, for the Dirichlet process, and remain fixed.

## 6. Analysis

In this section we provide further analysis of: (i) factors that influence the effectiveness of language pairings in bilingual models, (ii) the incremental value of adding more languages in the latent vari-able model, (iii) the superlingual tags and their corresponding cross-lingual patterns as learned by the latent variable model, and (iv) whether multilingual data is more helpful than additional monolingual data. We focus here on the full lexicon scenario, though we expect our analysis to extend to the various reduced lexicon cases considered above as well.

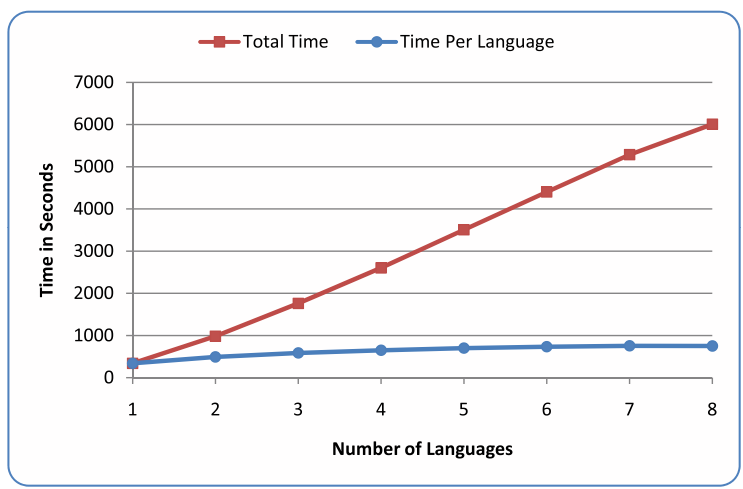

Figure 4: Average running time for 1000 iterations of the latent variable model. Results are averaged over all possible language subsets of each size. The top line shows the average running time for the entire subset, and the bottom line shows the running time divided by the number of languages.

### 6.1 Predicting Effective Language Pairings

We first analyze the cross-lingual variation in performance for different bilingual language pairings. As shown in Table 10, the performance of the merged node model for each target language varies substantially across pairings. In addition, the identity of the optimally helpful language pairing also depends heavily on the target language in question. For instance, Slovene, achieves a large improvement when paired with Serbian (+7.4), a closely related Slavic language, but only a minor improvement when coupled with English (+1.8). On the other hand, for Bulgarian, the best performance is achieved when coupling with English (+6) rather than with closely related Slavic languages (+2.4 and +0). Thus, optimal pairings do not correspond simply to language relatedness. We note that when applying multilingual learning to morphological segmentation the best results were obtained for related languages, but only after incorporating declarative knowledge about their lower-level phonological relations using a prior which encourages phonologically close aligned morphemes (Snyder & Barzilay, 2008). Here too, a more complex model which models lower-level morphological relatedness (such as case) may yield better outcomes for closely related languages.

As an upper bound on the merged node model performance, line 7 of Table 10 shows the results when selecting (with the help of an oracle) the best partner for each language. The average accuracy using this oracle is 95.4%, substantially higher than the average bilingual pairing accuracy of 93.2%, and even somewhat higher than the latent variable model performance of 95%. This gap in performance motivates a closer examination of the relationship between languages that constitute effective pairings.

| | | | | | coupled with... | | | | |
|---|---|---|---|---|---|---|---|---|---|
| | **Avg** | BG | CS | EN | ET | HU | RO | SL | SR |
| BG | 91.3 | | 90.2 | **94.7** | 92.3 | 90.6 | 91.2 | 91.1 | 88.7† |
| CS | 96.9 | 95.3 | | 97.5 | **97.8** | 96.3 | 96.4 | 97.4 | 97.4 |
| EN | 95.9 | **96.1** | 95.9† | | 95.8† | 95.8† | 95.8† | **96.1** | 96.0 |
| ET | 93.3 | 93.0 | 94.0 | 92.9† | | 92.2† | 93.0 | **94.2** | 93.9 |
| HU | 96.7 | 96.8 | 96.6 | 96.8 | **96.9** | | 96.8 | 96.5 | 96.7 |
| RO | 91.9 | **94.1** | 90.6† | 92.0 | 91.3 | 90.3† | | 91.3 | 93.9 |
| SL | 89.3 | 88.5 | 88.1 | 89.2 | 89.8 | 87.5† | 87.5† | | **94.8** |
| SR | 90.2 | 88.5 | 88.2 | **94.5** | 94.2 | 89.5 | 85.0 | 91.4 | |

*(Table header: MERGEDNODE MODEL; left axis label: accuracy for...)*

Table 10: Merged node model accuracy for all language pairs. Each row corresponds to the performance of one language, each column indicates the language with which the performance was achieved. The best result for each language is indicated in bold. All results other than those marked with a † are significantly higher than the monolingual baseline at $p < 0.05$ according to a sign test.

### 6.1.1 CROSS-LINGUAL ENTROPY

In a previous publication (Snyder et al., 2008) we proposed using *cross-lingual entropy* as a post-hoc explanation for variation in coupling performance. This measure calculates the entropy of a tagging decision in one language given the identity of an aligned tag in the other language. While cross-lingual entropy seemed to correlate well with relative performance for the four languages considered in that publication, we find that it does not correlate as strongly for all eight languages considered here. We computed the Pearson correlation coefficient (Myers & Well, 2002) between the relative bilingual performance and cross-lingual entropy. For each target language, we rank the remaining seven languages based on two measures: how well the paired language contributes to improved performance of the target, and the cross-lingual entropy of the target language given the coupled language. We compute the Pearson correlation coefficient between these two rankings to assess their degree of overlap. See Table 19 in the Appendix for a complete list of results. On average, the coefficient was 0.29, indicating a weak positive correlation between relative bilingual performance and cross-lingual entropy.

### 6.1.2 ALIGNMENT DENSITY

We note that even if cross-lingual entropy had exhibited higher correlation with performance, it would be of little practical utility in an unsupervised scenario since its estimation requires a tagged corpus. Next we consider the density of pairwise lexical alignments between language pairs as a predictive measure of their coupled performance. Since alignments constitute the multilingual anchors of our models, as a practical matter greater alignment density should yield greater opportunities for cross-lingual transfer. From the linguistic viewpoint, this measure may also indirectly capture the correspondence between two languages. Moreover, this measure has the benefit of being computable from an untagged corpus, using automatically obtained GIZA++ alignments. As before, for each target language, we rank the other languages by relative bilingual performance, as well as by the percentage of words in the target language to which they provide alignments. Here we find an average Pearson coefficient of 0.42, indicating mild positive correlation. In fact, if we use alignment density as a criterion for selecting optimal pairing decisions for each target language, we obtain an average accuracy of 94.67% — higher than average bilingual performance, but still somewhat below the performance of the multilingual model.

### 6.1.3 MODEL CHOICE

The choice of model may also contribute to the patterns of variability we observe across language pairs. To test this hypothesis, we ran our latent variable model on all pairs of languages. The results of this experiment are shown in Table 11. As in the case of the merged node model, the performance of each target language depends heavily on the choice of partner. However, the exact patterns of variability differ in this case from those observed for the merged node model. To measure this variability, we compare the pairing preferences for each language under each of the two models. More specifically, for each target language we rank the remaining seven languages by their contribution under each of our two models, and compute the Pearson coefficient between these two rankings. As seen in the last column of Table 19 in the Appendix, we find a coefficient of 0.49 between the two rankings, indicating positive, though far from perfect, correlation.

| | | LATENTVARIABLE MODEL | | | | | | | |
|---|---|---|---|---|---|---|---|---|---|
| | | *coupled with...* | | | | | | | |
| | **Avg** | BG | CS | EN | ET | HU | RO | SL | SR |
| BG | 91.9 | | 92.2 | 91.9 | 91.6 | 91.6 | 92.1 | **92.3** | 91.8 |
| CS | 97.2 | 97.5 | | 97.5 | **97.6** | 97.4 | 97.4 | 96.5 | 96.8 |
| EN | 95.7 | 95.7† | 95.7† | | 95.7† | 95.6† | 95.7† | 95.7† | **95.8†** |
| ET | 93.9 | **94.8** | 94.3 | 93.4 | | 92.3† | 93.9 | 94.5 | 94.1 |
| HU | 96.8 | **97.0** | 96.8 | 96.7 | 96.7 | | 96.8 | 96.6 | 96.8 |
| RO | 93.2 | 94.6 | 92.1 | 92.4 | 92.3 | 92.1 | | 94.4 | **94.7** |
| SL | 90.5 | 88.6 | 87.7 | 92.4 | **95.2** | 87.5† | 87.6† | | 94.6 |
| SR | 91.6 | **94.7** | 88.5 | 94.5 | 94.5 | 89.7 | 88.0 | 91.1 | |

*accuracy for...*

Table 11: Accuracy of latent variable model when run on language pairs. Each row corresponds to the performance of one language, each column indicates the language with which the performance was achieved. The best result for each language is indicated in bold. All results other than those marked with a † are significantly higher than the monolingual baseline at $p < 0.05$ according to a sign test.

### 6.1.4 UTILITY OF EACH LANGUAGE AS A BILINGUAL PARTNER

We also analyze the overall *helpfulness* of each language. As before, for each target language, we rank the remaining seven languages by the degree to which they contribute to increased target language performance when paired in a bilingual model. We can then ask whether the helpfulness rankings provided by each of the eight languages are correlated with one another — in other words, whether languages tend to be universally helpful (or unhelpful) or whether helpfulness depends heavily on the identity of the target language. We consider all pairs of target languages, and compute the Pearson rank correlation between their rankings of the six supplementary languages that they have in common (excluding the two target languages themselves). When we average these pairwise rank correlations we obtain a coefficient of 0.20 for the merged node model and 0.21 for the latent variable model. These low correlations indicate that language helpfulness depends crucially on the target language in question. Nevertheless, we can still compute the average helpfulness of each language (across all target languages) to obtain something like a "universal" helpfulness ranking. See Table 20 in the appendix for this ranking. We can then ask whether this ranking correlates with language properties which might be predictive of general helpfulness. We compare the universal helpfulness rankings[10] to language rankings induced by tag-per-token ambiguity (the average number of tags allowed by the dictionary per token in the corpus) as well as trigram entropy (the entropy of the tag distribution given the previous two tags). In both cases we assign the highest rank to the language with *lowest* value, as we expect lower entropy and ambiguity to correlate with greater helpfulness. Contrary to expectations, the ranking induced by tag-per-token ambiguity actually correlates *negatively* with both universal helpfulness rankings by very small amounts (-0.28 for the merged node model and -0.23 for the latent variable model). For both models, Hungarian, which has the lowest tag-per-token ambiguity of all eight languages, had the *worst* universal helpfulness ranking. The correlations with trigram entropy were only a little more predictable. In the case of the latent variable model, there was no correlation at all between trigram entropy and universal helpfulness (-0.01). In the case of the merged node model, however, there was moderate positive correlation (0.43).

## 6.2 Adding Languages in the Latent Variable Model

While bilingual performance depends heavily on the choice of language pair, the latent variable model can easily incorporate all available languages, obviating the need for any choice. To test performance as the number of languages increases, we ran the latent variable model with all possible subsets of the eight languages in the full lexicon as well as all three reduced lexicon scenarios. Figures 5, 6, 7, and 8 plot the average accuracy as the number of available languages varies for all four lexicon scenarios (in decreasing order of the lexicon size). For comparison, the monolingual and average bilingual baseline results are given. In all scenarios, our latent variable model steadily gains in accuracy as the number of available languages increases, and in most scenarios sees an appreciable uptick when going from seven to eight languages. In the full lexicon case, the gap between supervised and unsupervised performance is cut by nearly two thirds under the unsupervised latent variable model with all eight languages.

Interestingly, as the lexicon is reduced in size, the performance of the bilingual merged node model gains relative to the latent variable model on pairs. In the full lexicon case, the latent variable model is clearly superior, whereas in the two moderately reduced lexicon cases, the performance on pairs is more or less the same for the two models. In the case of the drastically reduced lexicon

---

10. We note that the universal helpfulness rankings obtained from each of the two multilingual models match each other only roughly: their correlation coefficient with one another is 0.50. In addition, "universal" in this context refers only to the eight languages under consideration and the rankings could very well change in a wider multilingual context.

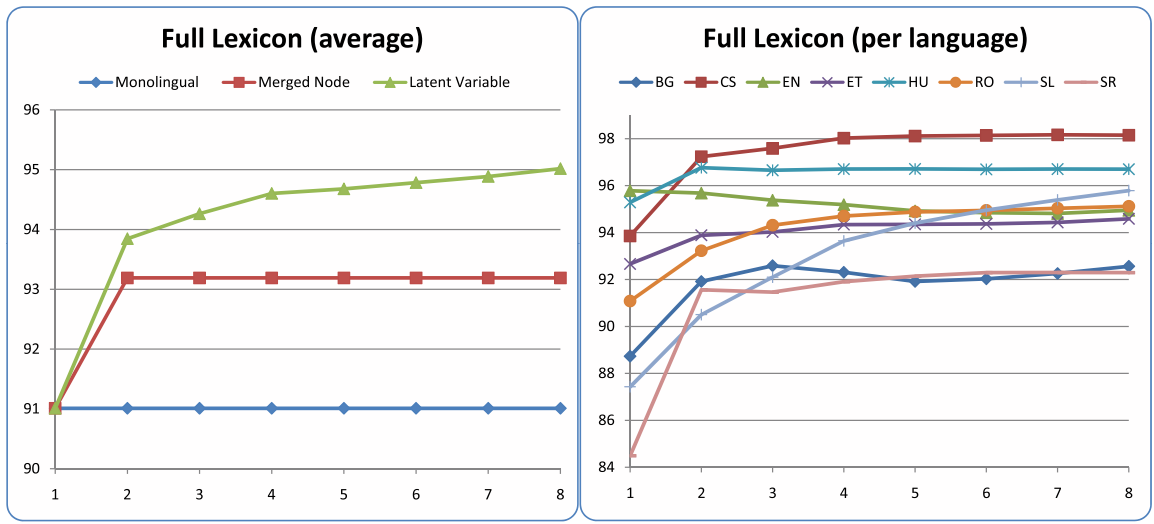

Figure 5: The performance of the latent variable model as the number of languages varies (averaged over all subsets of the eight languages for each size). LEFT: Average performance across all languages. Scores for monolingual and bilingual merged node models are given for comparison. RIGHT: The Performance for each individual language as the number of available languages varies.

(100 words), the merged node model is the clear winner. Thus, it seems that of the two models, the performance gains of the latent variable model are more sensitive to the size of the lexicon.

The same four figures (5, 6, 7, and 8) also show the multilingual performance broken down by language. All languages except for English tend to increase in accuracy as additional languages are added to the mix. Indeed, in the two cases of moderately reduced lexicons (Figures 6 and 7) all languages except for English show steady large gains which actually increase in size when going from seven to the full set of eight languages. In the full lexicon case (Figure 5), Estonian, Romanian, and Slovene display steady increases until the very end. Hungarian peaks at two languages, Bulgarian at three languages, and Czech and Serbian at seven languages. In the more drastic reduced lexicon case (Figure 8), the performance across languages is less consistent and the gains when languages are added are less stable. All languages report gains when going from one to two languages, but only half of them increase steadily up to eight languages. Two languages seem to trend downward after two or three languages, and the other two show mixed behavior.

In the full lexicon case (Figure 5), English is the only language which fails to improve. In the other scenarios, English gains initially but these gains are partially eroded when more languages are added. It is possible that English is an outlier since it has significantly lower tag transition entropy than any of the other languages (see Table 3). Thus it may be that internal tag transitions are simply more informative for English than any information that can be gleaned from multilingual context.

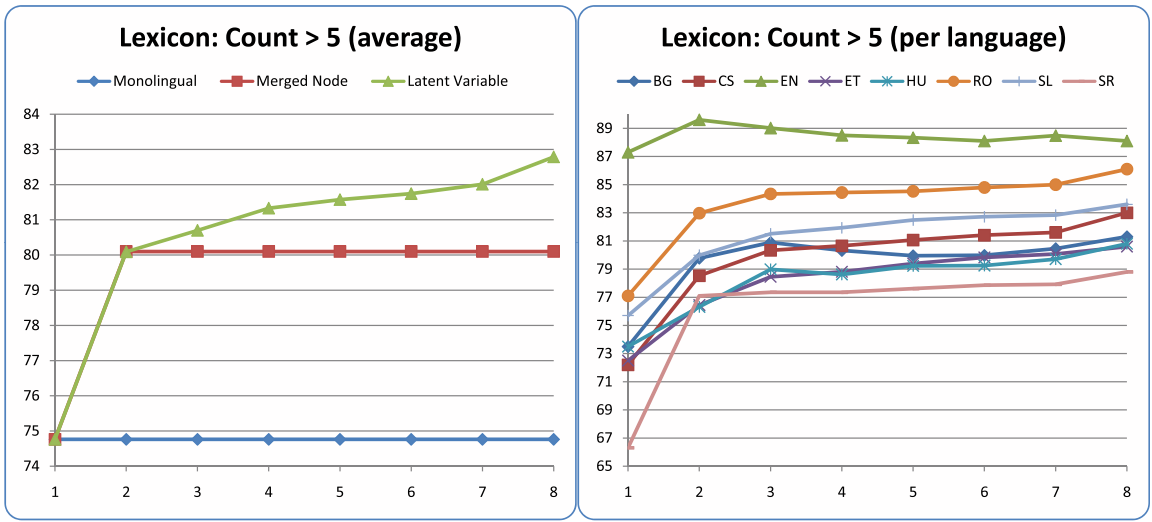

Figure 6: The performance of the latent variable model for the reduced lexicon scenario (Counts > 5), as the number of languages varies (averaged over all subsets of the eight languages for each size). LEFT: Average performance across all languages. Scores for monolingual and bilingual merged node models are given for comparison. RIGHT: The Performance for each individual language as the number of available languages varies.

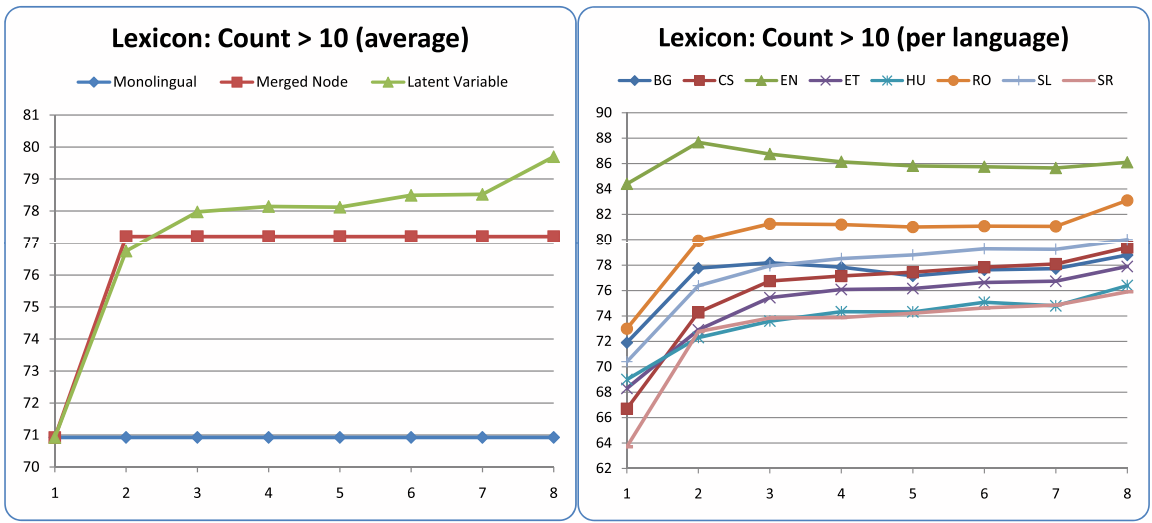

Figure 7: The performance of the latent variable model for the reduced lexicon scenario (Counts > 10), as the number of languages varies (averaged over all subsets of the eight languages for each size). LEFT: Average performance across all languages. Scores for monolingual and bilingual merged node models are given for comparison. RIGHT: The Performance for each individual language as the number of available languages varies.

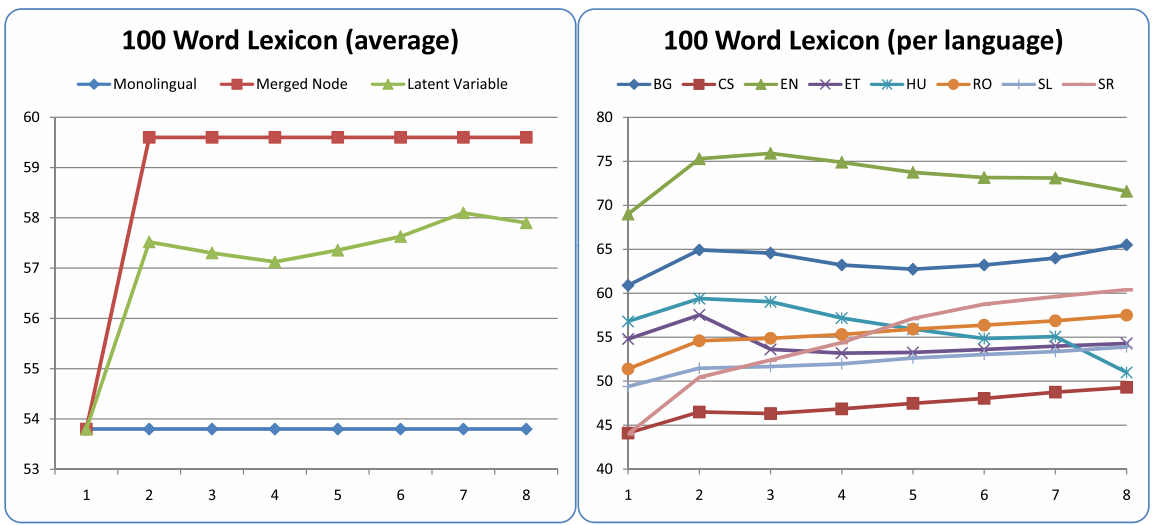

Figure 8: The performance of the latent variable model for the reduced lexicon scenario (100 words), as the number of languages varies (averaged over all subsets of the eight languages for each size). LEFT: Average performance across all languages. Scores for monolingual and bilingual merged node models are given for comparison. RIGHT: The Performance for each individual language as the number of available languages varies.

### 6.3 Analysis of the Superlingual Tag Values

In this section we analyze the superlingual tags and their corresponding part-of-speech distributions, as learned by the latent variable model. Recall that each superlingual tag intuitively represents a discovered *multilingual context* and that it is through these tags that multilingual information is propagated. More formally, each superlingual tag provides a complete distribution over parts-of-speech for each language, allowing the encoding of both primary and secondary preferences separately for each language. These preferences then interact with the language-specific context (i.e. the surrounding parts-of-speech and the corresponding word). We place a Dirichlet process prior on the superlingual tags, so the number of sampled values is dictated by the complexity of the data. In fact, as shown in Table 12, the number of sampled superlingual tags steadily increases with the number of languages. As multilingual contexts becomes more complex and diverse, additional superlingual tags are needed.

| Number of languages | 2 | 3 | 4 | 5 | 6 | 7 | 8 |
|---|---|---|---|---|---|---|---|
| Number of superlingual tag values | 11.07 | 12.57 | 13.87 | 15.07 | 15.79 | 16.13 | 16.50 |

Table 12: Average number of sampled superlingual tag values as the number of languages increases.

Next we analyze the part-of-speech tag distributions associated with superlingual tag values. Most superlingual tag values correspond to low entropy tag distributions, with a single dominant part-of-speech tag across all languages. See, for example, the distributions associated with superlin-gual tag value 6 in Table 13, all of which favor nouns by large margins. Similar sets of distributions occur favoring verbs, adjectives, and the other primary part-of-speech categories. In fact, among the seventeen sampled superlingual tag values, nine belong to this type, and they cover 80% of actual superlingual tag instances. The remaining superlingual tags correspond to more complex cross-lingual patterns. The associated tag distributions in those cases favor different part-of-speech tags in various languages and tend to have higher entropy, with the probability mass spread more evenly over two or three tags. One such example is the set of distributions associated with the superlingual tag value 14 in Table 13, which seems to be a mixed noun/verb class. In six out of eight languages the most favored tag is verb, while a strong secondary choice in these cases is noun. However, for Estonian and Hungarian, this preference is reversed, with nouns being given higher probability. This superlingual tag may have captured the phenomenon of "light verbs," whereby verbs in one language correspond to a combination of a noun and verb in another language. For example the English verb *whisper*/V, when translated into Urdu, becomes the collocation *whisper*/N *do*/V. In these cases, verbs and nouns will often be aligned to one another, requiring a more complex superlingual tag. The analysis of these examples shows that the superlingual tags effectively learns both simple and complex cross-lingual patterns

| | | | | | |
|---|---|---|---|---|---|
| | BG | $P(N) = 0.91,\ P(A) = 0.04,\ ...$ | | BG | $P(V) = 0.66,\ P(N) = 0.21,\ ...$ |
| | CS | $P(N) = 0.92,\ P(A) = 0.03,\ ...$ | | CS | $P(V) = 0.60,\ P(N) = 0.22,\ ...$ |
| | EN | $P(N) = 0.97,\ P(V) = 0.00,\ ...$ | | EN | $P(V) = 0.55,\ P(N) = 0.25,\ ...$ |
| TAG VALUE 6 | ET | $P(N) = 0.91,\ P(V) = 0.03,\ ...$ | TAG VALUE 14 | ET | $P(N) = 0.52,\ P(V) = 0.29,\ ...$ |
| | HU | $P(N) = 0.85,\ P(A) = 0.06,\ ...$ | | HU | $P(N) = 0.44,\ P(V) = 0.34,\ ...$ |
| | RO | $P(N) = 0.90,\ P(A) = 0.04,\ ...$ | | RO | $P(V) = 0.45,\ P(N) = 0.33,\ ...$ |
| | SL | $P(N) = 0.94,\ P(A) = 0.03,\ ...$ | | SL | $P(V) = 0.55,\ P(N) = 0.24,\ ...$ |
| | SR | $P(N) = 0.92,\ P(A) = 0.03,\ ...$ | | SR | $P(V) = 0.49,\ P(N) = 0.26,\ ...$ |

Table 13: Part-of-speech tag distributions associated with two superlingual latent tag values. Probabilities of only the two most probable tags for each language are shown.

### 6.3.1 PERFORMANCE WITH REDUCED DATA

One potential objection to the claims made in this section is that the improved results may be due merely to the addition of more data, so that the multilingual aspect of the model may be irrelevant. We test this idea by evaluating the monolingual, merged node, and latent variable systems on training sets in which the number of examples is reduced by half. The multilingual models in this setting have access to exactly half as much data as the monolingual model in the original experiment. As shown in Table 14, both the monolingual baseline and our models are quite insensitive to this drop in data. In fact, both of our models, when trained on half of the corpus, still outperform the monolingual model trained on the entire corpus. This indicates that the performance gains demonstrated by multilingual learning cannot be explained merely by the addition of more data.

|  | **Avg** | BG | CS | EN | ET | HU | RO | SL | SR |
|---|---|---|---|---|---|---|---|---|---|
| Monolingual: full data | 91.2 | 88.7 | 93.9 | 95.8 | 92.7 | 95.3 | 91.1 | 87.4 | 84.5 |
| Monolingual: half data | 91.0 | 88.8 | 93.8 | 95.7 | 92.6 | 95.3 | 90.2 | 87.5 | 84.5 |
| MergedNode: (*avg.*) full data | 93.2 | 91.3 | 96.9 | 95.9 | 93.3 | 96.7 | 91.9 | 89.3 | 90.2 |
| MergedNode: (*avg.*) half data | 93.0 | 91.1 | 96.6 | 95.7 | 92.7 | 96.7 | 92.0 | 88.9 | 89.9 |
| LatentVariable: full data | 95.0 | 92.6 | 98.2 | 95.0 | 94.6 | 96.7 | 95.1 | 95.8 | 92.3 |
| LatentVariable: half data | 94.7 | 92.6 | 97.8 | 94.7 | 93.9 | 96.7 | 94.4 | 95.4 | 92.2 |

Table 14: Tagging accuracy on reduced training dataset, with complete tag dictionaries; results on the full training dataset are repeated here for comparison. The first column reports average results across all languages (see Table 3 for language name abbreviations).

## 7. Conclusions

The key hypothesis of multilingual learning is that by combining cues from multiple languages, the structure of each becomes more apparent. We considered two ways of applying this intuition to the problem of unsupervised part-of-speech tagging: a model that directly merges tag structures for a pair of languages into a single sequence and a second model which instead incorporates multilingual context using latent variables.

Our results demonstrate that by incorporating multilingual evidence we can achieve impressive performance gains across a range of scenarios. When a full lexicon is available, our two models cut the gap between unsupervised and supervised performance by nearly one third (merged node model, averaged over all pairs) and two thirds (latent variable model, using all eight languages). For all but one language, we observe performance gains as additional languages are added. The sole exception is English, which only gains from additional languages in reduced lexicon settings.

In most scenarios, the latent variable model achieves better performance than the merged node model, and has the additional advantage of scaling gracefully with the number of languages. These observations suggest that the non-parametric latent variable structure provides a more flexible paradigm for incorporating multilingual cues. However, the benefit of the latent variable model relative to the merged node model (even when running both models on pairs of languages) seems to decrease with the size of the lexicon. Thus, in practical scenarios where only a small lexicon or no lexicon is available, the merged node model may represent a better choice.

Our experiments have shown that performance can vary greatly depending on the choice of additional languages. It is difficult to predict *a priori* which languages constitute good combinations. In particular, language relatedness itself cannot be used as a consistent predictor as sometimes closely related languages constitute beneficial couplings and sometimes unrelated languages are more helpful. We identify a number of features which correlate with bilingual performance, though we observe that these features interact in complex ways. Fortunately, our latent variable model allows us to bypass this question by simply using all available languages.

In both of our models, lexical alignments play a crucial role as they determine the typology of the model for each sentence. In fact, we observed a positive correlation between alignment density and bilingual performance, indicating the importance of high quality alignments. In our experiments, we considered the alignment structure an observed variable, produced by standard MT tools which operate over pairs of languages. An interesting alternative would be to incorporate alignment structure into the model itself, to find alignments best tuned for tagging accuracy based on the evidence of multiple languages rather than pairs.

Another limitation of the two models is that they only consider one-to-one lexical alignments. When pairing isolating and synthetic languages[11] it would be beneficial to align short analytical phrases consisting of multiple words to single morpheme-rich words in the other language. To do so would involve flexibly aligning and chunking the parallel sentences throughout the learning process.

An important direction for future work is to incorporate even more sources of multilingual information, such as additional languages and declarative knowledge of their typological properties (Comrie, 1989). In this paper we showed that performance improves as the number of languages increases. We were limited by our corpus to eight languages, but we envision future work on massively parallel corpora involving dozens of languages as well as learning from languages with non-parallel data.

## Bibliographic Note

Portions of this work were previously presented in two conference publications (Snyder, Naseem, Eisenstein, & Barzilay, 2008, 2009). The current article extends this work in several ways, most notably: we present a new inference procedure for the merged node model which yields improved results (Section 3.1.2) and we conduct extensive new empirical analyses of the multilingual results. More specifically, we analyze properties of language combinations that contribute to successful multilingual learning, we show that adding multilingual data provides much greater benefit than increasing the quantity of monolingual data, we investigate additional scenarios of lexicon reduction, we investigate scalability as a function of the number of languages, and we experiment with semi-supervised settings (Sections 5 and 6).

## Acknowledgments

The authors acknowledge the support of the National Science Foundation (CAREER grant IIS-0448168 and grants IIS-0835445, IIS-0904684) and the Microsoft Research Faculty Fellowship. Thanks to Michael Collins, Tommi Jaakkola, Amir Globerson, Fernando Pereira, Lillian Lee, Yoong Keok Lee, Maria Polinsky and the anonymous reviewers for helpful comments and suggestions. Any opinions, findings, and conclusions or recommendations expressed above are those of the authors and do not necessarily reflect the views of the NSF.

---

11. Isolating languages are those with a morpheme to word ratio close to one, and synthetic languages are those which allow multiple morphemes to be easily combined into single words. English is an example of an isolating language, whereas Hungarian is a synthetic language.

## Appendix A. Alignment Statistics

|    | BG | CS | EN | ET | HU | RO | SL | SR |
|----|----|----|----|----|----|----|----|----|
| BG |    | 42163 | 51098 | 33849 | 31673 | 42017 | 45969 | 46434 |
| CS | 42163 |    | 43067 | 40207 | 31537 | 32559 | 57789 | 49740 |
| EN | 51098 | 43067 |    | 40746 | 39012 | 50289 | 52869 | 48394 |
| ET | 33849 | 40207 | 40746 |    | 32056 | 27709 | 42499 | 37681 |
| HU | 31673 | 31537 | 39012 | 32056 |    | 26455 | 34072 | 29797 |
| RO | 42017 | 32559 | 50289 | 27709 | 26455 |    | 36442 | 38004 |
| SL | 45969 | 57789 | 52869 | 42499 | 34072 | 36442 |    | 59865 |
| SR | 46434 | 49740 | 48394 | 37681 | 29797 | 38004 | 59865 |    |

Table 15: Number of alignments per language pair

|    | BG | CS | EN | ET | HU | RO | SL | SR | Avg. |
|----|----|----|----|----|----|----|----|----|------|
| BG |    | 2.77 | 6.13 | 3.36 | 4.04 | 4.52 | 2.95 | 3.48 | 3.89 |
| CS | 2.77 |    | 3.67 | 1.92 | 2.73 | 3.61 | 2.59 | 2.64 | 2.85 |
| EN | 6.13 | 3.67 |    | 4.35 | 6.12 | 5.59 | 3.54 | 3.86 | 4.75 |
| ET | 3.36 | 1.92 | 4.35 |    | 2.88 | 3.88 | 2.44 | 2.21 | 3.01 |
| HU | 4.04 | 2.73 | 6.12 | 2.88 |    | 4.13 | 3.09 | 3.06 | 3.72 |
| RO | 4.52 | 3.61 | 5.59 | 3.88 | 4.13 |    | 3.78 | 3.92 | 4.20 |
| SL | 2.95 | 2.59 | 3.54 | 2.44 | 3.09 | 3.78 |    | 4.11 | 3.22 |
| SR | 3.48 | 2.64 | 3.86 | 2.21 | 3.06 | 3.92 | 4.11 |    | 3.33 |

Table 16: Percentage of alignments removed per language pair

## Appendix B. Tag Repository

|  | BG | CS | EN | ET | HU | RO | SL | SR |
|---|---|---|---|---|---|---|---|---|
| Adjective | x | x | x | x | x | x | x | x |
| Conjunction | x | x | x | x | x | x | x | x |
| Determiner | - | - | x | - | - | x | - | - |
| Interjection | x | x | x | x | x | x | x | x |
| Numeral | x | x | x | x | x | x | x | x |
| Noun | x | x | x | x | x | x | x | x |
| Pronoun | x | x | x | x | x | x | x | x |
| Particle | x | x | - | - | - | x | x | x |
| Adverb | x | x | x | x | x | x | x | x |
| Adposition | x | x | x | x | x | x | x | x |
| Article | - | - | - | - | x | x | - | - |
| Verb | x | x | x | x | x | x | x | x |
| Residual | x | x | x | x | x | x | x | x |
| Abbreviation | x | x | x | x | x | x | x | x |

Table 17: Tag repository for each language

## Appendix C. Stanford Tagger Performance

| Language | Accuracy |
|---|---|
| BG | 96.1 |
| CS | 97.2 |
| EN | 97.6 |
| ET | 97.1 |
| HU | 96.3 |
| RO | 97.6 |
| SL | 96.6 |
| SR | 95.5 |
| Avg. | 96.7 |

Table 18: Performance of the (supervised) Stanford tagger for the full lexicon scenario

## Appendix D. Rank Correlation

| Performance correlates for MERGEDNODE model | | | |
|---|---|---|---|
| Language | Cross-lingual entropy | Alignment density | LATENTVARIABLE performance |
| BG | -0.29 | 0.09 | -0.09 |
| CS | 0.39 | 0.34 | 0.24 |
| EN | 0.28 | 0.77 | 0.42 |
| ET | 0.46 | 0.56 | 0.56 |
| HU | 0.31 | -0.02 | 0.29 |
| RO | 0.34 | 0.83 | 0.89 |
| SL | 0.59 | 0.66 | 0.95 |
| SR | 0.21 | 0.13 | 0.63 |
| Avg. | 0.29 | 0.42 | 0.49 |
| Performance correlates for LATENTVARIABLE model | | | |
| Language | Cross-lingual entropy | Alignment density | MERGEDNODE performance |
| BG | 0.58 | 0.44 | -0.09 |
| CS | -0.40 | -0.44 | 0.24 |
| EN | 0.67 | 0.41 | 0.42 |
| ET | 0.14 | 0.32 | 0.56 |
| HU | -0.14 | -0.72 | 0.29 |
| RO | 0.04 | 0.68 | 0.89 |
| SL | 0.57 | 0.54 | 0.95 |
| SR | 0.18 | 0.10 | 0.68 |
| Avg. | 0.21 | 0.17 | 0.49 |

Table 19: Pearson correlation coefficients between bilingual performance on the target language and various rankings of the supplementary language. For both models and for each target language, we obtain a ranking over all supplementary languages based on bilingual performance in the target language. These rankings are then correlated with other characteristics of the bilingual pairing: **cross-lingual entropy** (the entropy of tag distributions in the target language given aligned tags in the supplementary language); **alignment density** (the percentage of words in the target language aligned to words in the auxiliary language); and performance in the alternative model (target language performance when paired with the same supplementary language in the alternative model).

## Appendix E. Universal Helpfulness

| MERGEDNODE model | | LATENTVARIABLE model | |
|---|---|---|---|
| ET | 2.43 | BG | 1.86 |
| EN | 2.57 | SR | 3.00 |
| SL | 3.14 | ET | 3.14 |
| BG | 3.43 | CS | 3.71 |
| SR | 3.43 | EN | 3.71 |
| RO | 4.71 | SL | 3.71 |
| CS | 5.00 | RO | 4.14 |
| HU | 5.71 | HU | 6.00 |

Table 20: Average helpfulness rank for each language under the two models